\documentclass[lettersize,journal]{IEEEtran}
\usepackage{amsmath,amsfonts}
\usepackage{algorithmic}
\usepackage{algorithm}
\usepackage{array}
\usepackage[caption=false,font=normalsize,labelfont=sf,textfont=sf]{subfig}
\usepackage{textcomp}
\usepackage{stfloats}
\usepackage{url}
\usepackage{verbatim}
\usepackage{graphicx}

\usepackage{cite}
\usepackage{hyperref}
\newcommand{\ie}{\emph{i.e.}}
\newcommand{\eg}{\emph{e.g.}} 
\usepackage{xcolor}         
\usepackage{booktabs, multirow}
\usepackage[table]{xcolor}
\definecolor{Gray}{gray}{0.92}
\usepackage{amssymb}   %
\usepackage{wasysym}   %
\usepackage[normalem]{ulem}
\usepackage{enumitem} 
\usepackage[dvipsnames]{xcolor}
\definecolor{Gray}{gray}{0.9}

\begin{document}

\title{Echo Planning for Autonomous Driving: From Current \\ Observations to Future Trajectories and Back}

\author{Jintao Sun, Hu Zhang, Gangyi Ding, Zhedong Zheng
\thanks{T. Sun and G. Ding are with the School of Computer Science and Technology, Beijing Institute of Technology, China 100081. E-mail: 3120215524@bit.edu.cn, dgy@bit.edu.cn}
\thanks{H. Zhang is with CSIRO DATA61, Australia. E-mail: hu1.zhang@csiro.au}
\thanks{Z. Zheng is with the Faculty of Science and Technology and Institute of Collaborative Innovation, University of Macau, Macau SAR, China 999078. E-mail: zhedongzheng@um.edu.mo}}
\markboth{Journal of \LaTeX\ Class Files,~Vol.~18, No.~9, September~2020}%
{Shell \MakeLowercase{\textit{et al.}}: A Sample Article Using IEEEtran.cls for IEEE Journals}


\maketitle

\begin{abstract}
Modern end-to-end autonomous driving systems suffer from a critical limitation: their planners lack mechanisms to enforce temporal consistency between predicted trajectories and evolving scene dynamics. This absence of self-supervision allows early prediction errors to compound catastrophically over time. We introduce Echo Planning (\textbf{EchoP}), a new self-correcting framework that establishes an end-to-end Current → Future → Current (CFC) cycle to harmonize trajectory prediction with scene coherence. Our key insight is that plausible future trajectories should be bi-directionally consistent, \ie, not only generated from current observations but also capable of reconstructing them. The CFC mechanism first predicts future trajectories from the Bird’s-Eye-View (BEV) scene representation, then inversely maps these trajectories back to estimate the current BEV state. By enforcing consistency between the original and reconstructed BEV representations through a cycle loss, the framework intrinsically penalizes physically implausible or misaligned trajectories. Experiments on nuScenes show that the proposed method yields competitive performance, reducing L2 error (Avg) by -0.04 m and collision rate by -0.12\% compared to one-shot planners. Moreover, EchoP seamlessly extends to closed-loop evaluation, \ie, Bench2Drive, attaining a 26.54\% success rate. Notably, EchoP requires no additional supervision: the CFC cycle acts as an inductive bias that stabilizes long-horizon planning. Overall, EchoP offers a simple, deployable pathway to improve reliability in safety-critical autonomous driving.
\end{abstract}


\begin{figure*}[htb]
  \centering
  \vspace{-.2in}
  \includegraphics[width=0.90\linewidth]{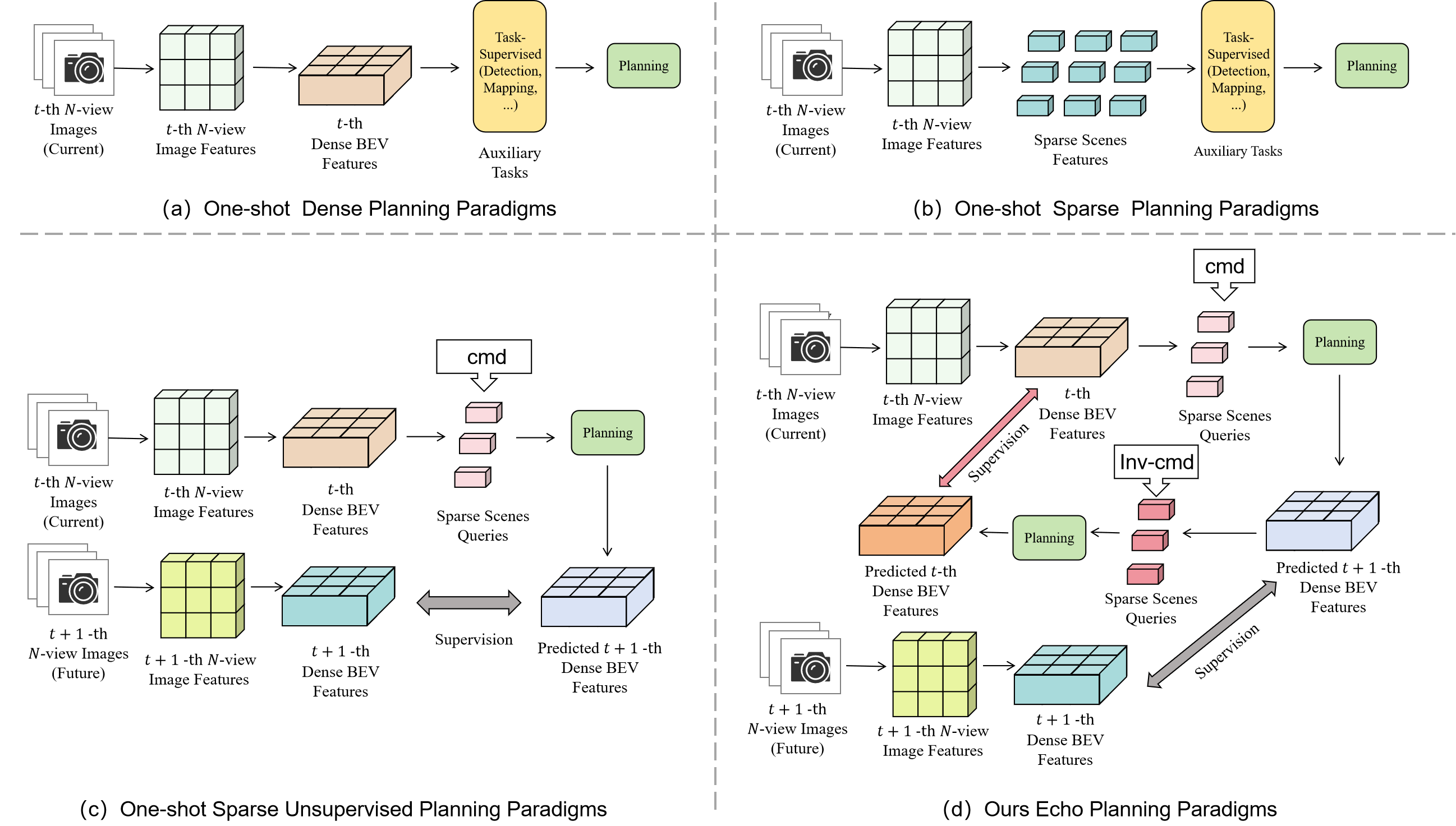}
    \vspace{-.1in}
  \caption{ \textbf{Comparison between one-shot paradigms and our echo planning paradigms.} (a), (b), and (c) represent different types of one-shot approaches. Both (a) and (b) construct scene representations from image features and rely on auxiliary tasks to supervise the model, thereby enhancing environmental understanding; their main distinction lies in whether dense BEV features are used. Method (c) discards the auxiliary tasks introduced in (b) and is among the first to highlight the importance of temporal supervision, though it still considers only forward verification and thus remains within the one-shot paradigm. In contrast, ours (d) presents the echo planning approach, which employs a \textbf{Current $\rightarrow$ Future $\rightarrow$ Current} cycle. This design enforces \textbf{bidirectional self-supervision}, allowing the model to validate scene understanding without additional auxiliary tasks.
  }
  \label{fig:1}
  \vspace{-.1in}
\end{figure*}

\section{Introduction}\label{1}
\IEEEPARstart{V}{ision}-based end-to-end planning has become a leading paradigm in autonomous driving research~\cite{hu2022st,hu2023planning,jiang2023vad,zheng2024genad,sun2024sparsedrive,weng2024drive,linavigation}.
Synchronized multi-view RGB images are usually mapped to Bird's-Eye-View (BEV) scene representation, 
 providing a direct interface between perception and planning. 
Recent comprehensive surveys~\cite{10.1109/TPAMI.2024.3483273,feng2025survey} have highlighted the critical role of progressive BEV perception in safety-critical driving.
\textcolor{black}{BEV perception can be broadly categorized into dense and sparse paradigms based on whether they adopt a dense grid representation.}
For instance, dense BEV models rasterize the scene into grids and learn rich spatial features~\cite{chen2024ppad,hu2022st,hu2023planning,jiang2023vad,zheng2024genad,ye2023fusionad,liu2023bevfusion,zhang2024graphad}. Advanced methods have further extended this to precise 3D lane detection~\cite{10.1109/TPAMI.2024.3508798}, bounding boxes~\cite{zhang2024approaching,zhang2025harnessing}, and vectorized road network translation~\cite{11175545}, providing explicit geometric constraints for downstream tasks. In contrast, sparse BEV models replace the grid with compact scene queries that reduce computation and label burden while preserving BEV geometry~\cite{sun2024sparsedrive,zhu2024autonomous,linavigation,zhang2024sparsead}. 
Based on the BEV representation, most current approaches follow a one-shot planning paradigm, forecasting the vehicle's future path solely from a single snapshot of the present scene, shown in Fig.~\ref{fig:1} (a, b, c). \textcolor{black}{As summarized in Fig.~\ref{fig:1}(a--c), existing one-shot end-to-end planning paradigms can be grouped into three representative categories. 
Fig.~\ref{fig:1}(a) illustrates dense grid based BEV planning pipelines, where a rasterized BEV feature map is supervised by auxiliary tasks and then used for trajectory prediction, as exemplified by ST-P3~\cite{hu2022st}, UniAD~\cite{hu2023planning}, VAD~\cite{jiang2023vad}, and GenAD~\cite{zheng2024genad}. 
Fig.~\ref{fig:1}(b) illustrates one-shot sparse query based planning that replaces dense grids with compact scene queries while still relying on auxiliary supervision, as in SparseDrive~\cite{sun2024sparsedrive} and SparseAD~\cite{zhang2024sparsead}. 
Fig.~\ref{fig:1}(c) illustrates sparse planning without perception supervision that introduces temporal supervision via forward consistency, as represented by SSR~\cite{linavigation}. 
In contrast, Fig.~\ref{fig:1}(d) summarizes EchoP with a \textbf{Current$\rightarrow$Future$\rightarrow$Current (CFC)} formulation, where the predicted future BEV is further used to reconstruct the current BEV state. 
This backward (echo) reconstruction provides an explicit reverse-consistency signal that complements the forward temporal supervision and makes the system functionality in Fig.~\ref{fig:1}(d) directly aligned with the two-loop design in Fig.~\ref{fig:2}.}
However, these methods lack an internal mechanism to enforce temporal consistency.
\textcolor{black}{
For instance, while MomAD~\cite{song2025don} has improved inter step trajectory stability by smoothing consecutive predictions, most one shot planners still lack an intrinsic self verification mechanism that checks whether the predicted future scene evolution remains consistent with the initial BEV observation.
As a result, trajectory and scene self consistency verification through a bidirectional current future current check has remained underexplored.}
Without the temporal verification, early inaccuracies can accumulate over time and lead to unsafe driving behaviors.
Hence, an open research question persists: \textbf{\textit{how can we inherently embed self correcting scene trajectory consistency in trajectory planning without incurring additional annotation overhead or heuristic complexity?}}

To address this challenge, we introduce \textbf{EchoP}, a novel self-correcting trajectory planning framework that enforces intrinsic temporal coherence through an end-to-end CFC feedback cycle (see Fig.~\ref{fig:1} (d)). While prior work~\cite{linavigation} models temporal variation by predicting a future BEV map from the current BEV and planned trajectory, they only enforce forward consistency and neglect the reverse constraint. In contrast, the core intuition of our EchoP is that a robust trajectory should be bidirectionally consistent: it should not only represent a plausible future evolution of the current scene but also be able to accurately reconstruct the current scene when projected backward.
Inspired by the video generation~\cite{suo2024jointly}, our model first generates future trajectories using a BEV representation of the current scene, and subsequently employs an inverse reconstruction step to predict back the current BEV state from these anticipated trajectories. This CFC cycle 
penalizes trajectories that are physically implausible or inconsistent with the initial scene configuration, thereby ensuring bidirectional consistency.

By enforcing the CFC feedback cycle, our method systematically reduces the impact of evolving scene dynamics on planning, preserves temporal consistency in the driving context, and consequently boosts trajectory accuracy. Experiments on the widely used nuScenes dataset~\cite{caesar2020nuscenes} confirm that our approach achieves competitive performance, sharply lowering both L2 trajectory error and collision rates relative to existing one-shot planners. Crucially, these gains are obtained without any additional supervision or handcrafted heuristics, making the method both reliable and readily deployable. EchoP therefore provides a systematic method to intrinsically enforce trajectory–scene consistency, advancing end-to-end autonomous driving toward safer and more dependable operation. Our primary contributions are:
\begin{itemize}[leftmargin=*]
\item \textbf{Self-Correcting Planning.}
We propose a novel echo planning paradigm for autonomous driving, built upon a CFC feedback loop. Unlike one-shot prediction methods, we leverage predicted future trajectories to inversely reconstruct the current Bird’s-Eye-View (BEV) scene state. This reconstruction process enables implicit self-correction by penalizing implausible trajectories without requiring any external supervision.

\item \textbf{High-Fidelity Driving Performance.} Extensive experiments on open-loop driving benchmarks, specifically nuScenes, validate that EchoP arrives at a competitive average collision rate of 0.17 without additional inference overhead. Moreover, EchoP can be seamlessly extended to closed-loop trajectory prediction, achieving a 26.54 success rate on the Bench2Drive benchmark.
\end{itemize}

Furthermore, distinct from conventional planners that solely prioritize trajectory regression, EchoP introduces a critical spatio-temporal consistency regularizer into the learning process. By enforcing that predicted future trajectories should be capable of inversely reconstructing current observations, the proposed CFC mechanism serves as an intrinsic self-supervisory signal. This bidirectional validation significantly enhances the system's capability to learn robust visual representations, reducing physically implausible predictions arising from sensor noise or ambiguous scene dynamics, effectively acting as a logic coherence check within the planning loop.


\textcolor{black}{
\textbf{Novelty, significance, and practical implications.}
While one-shot end-to-end planning has been extensively studied, existing temporal-supervision designs are mostly feed-forward and lack an explicit mechanism to verify whether the predicted future remains consistent with the current observation.
EchoP provides a distinct insight by introducing \emph{bidirectional} trajectory--scene consistency via a Current$\rightarrow$Future$\rightarrow$Current (CFC) feedback constraint, which penalizes plans that cannot coherently reconstruct the present state.
This design improves practical adaptability because the CFC term serves as a \emph{training-time} self-supervision signal and does not change the test-time planning pathway.
Our findings suggest a lightweight principle for stakeholders and the research community: using bidirectional consistency as an intrinsic regularizer to enhance planning robustness, and using reconstruction-based consistency as a diagnostic signal for spatio-temporal reasoning in complex traffic.}

The remainder of this paper is organized as follows. Section~\ref{2} reviews the related literature regarding end-to-end autonomous driving and BEV representation learning. Section~\ref{3} elaborates on the proposed Echo Planning framework, providing a detailed formulation of the self-correcting CFC mechanism. Section~\ref{4} presents extensive quantitative and qualitative experiments on standard benchmarks, along with comprehensive ablation studies to validate the effectiveness of our approach. Finally, Section~\ref{5} concludes the paper and outlines potential directions for future research.

\section{Related work}\label{2}
\textbf{End-to-end planning.} 
Planning is the ultimate objective of the first phase of end-to-end autonomous driving. Early work relied on relatively simple neural networks that ignored much of the scene context and therefore offered limited interpretability~\cite{Pomerleau_1988,Codevilla_Muller_Lopez_Koltun_Dosovitskiy_2018,cheng2022mpnp,Bojarski_Testa_Dworakowski_Firner_Flepp_Goyal_Jackel_Monfort_Muller_Zhang_etal,codevilla2019exploring,Prakash_Chitta_Geiger_2021,dauner2023parting}. 
With the advent of large-scale datasets and stronger BEV perception, many learning-based planners have been proposed~\cite{hu2022st,wu2022trajectory,hu2023planning,jiang2023vad,zheng2024genad,jia2024bench,jia2023think,weng2024drive,sun2024sparsedrive,li2024ego,zhang2024sparsead,zhang2024graphad}. 
Most recent systems still follow a perception, prediction, and planning pipeline to maintain transparency. 
For instance, ST-P3~\cite{hu2022st} chains map perception, BEV occupancy prediction, and trajectory planning to infer future ego motion from surround cameras. 
UniAD~\cite{hu2023planning} introduces a unified query design that integrates detection, mapping, and motion forecasting. 
VAD~\cite{jiang2023vad} uses a vectorized scene representation to couple scene understanding with planning constraints, and 
GenAD~\cite{zheng2024genad} generates future trajectories for both the ego vehicle and other agents within a learned probabilistic latent space. 
A newer line of research skips the explicit perception and prediction stages to facilitate efficiency. 
BEV-Planner~\cite{li2024ego} employs ego queries to extract task-relevant cues directly from BEV features, while 
SSR~\cite{linavigation} leverages navigation commands to focus on salient regions without perception supervision. 
\textcolor{black}{We also note several very recent end-to-end planning directions that complement the above paradigms, including constraint-guided flow-based planning (GuideFlow~\cite{guideflow2025}), world-model and vision-language assisted planning (MindDrive~\cite{minddrive2025}), and probabilistic planning to model uncertainty (VADv2~\cite{vadv22024}).} 
\textcolor{black}{Despite the progress, many end-to-end planners remain predominantly feed-forward, and fewer works explicitly incorporate an intrinsic self-verification signal that checks whether predicted futures stay semantically consistent with the current scene representation, beyond trajectory-sequence regularization.} 
We address this gap with an end-to-end CFC cycle that enforces coherent scene evolution over time.

\textbf{BEV perception in autonomous driving.} 
Perception forms the bedrock of autonomous driving, as we need distill actionable information from raw sensor streams. Precise and efficient scene understanding is therefore essential. 
The steady evolution of BEV representations~\cite{chen2022efficient,Hu_Murez_Mohan_Dudas_Hawke_Badrinarayanan_Cipolla_Kendall_2021,huang2021bevdet,li2023bevdepth,liao2024lane,liao2022maptr,shao2023reasonnet,liu2023vision,zhang2022beverse,ju2025video2bev} has propelled perception forward, while the lower cost of RGB cameras is gradually supplanting LiDAR-based pipelines~\cite{chen2023voxelnext,Graham_Engelcke_Maaten_2018,Mao_Xue_Niu_Bai_Feng_Liang_Xu_Xu_2021,ye2023fusionad,yuan2024drama}. 
LSS~\cite{Philion_Fidler_2020} is a landmark effort that used depth prediction to lift perspective features into BEV space. 
To enhance feature robustness, depth-induced multi-scale attention mechanisms~\cite{9729103} have been explored to better capture saliency in complex depth ranges. 
BEVFormer~\cite{10791908} introduces spatial and temporal attention within a Transformer, achieving strong camera-only detection. 
Furthermore, DeepInteraction~\cite{10980037} demonstrates that predictive interaction between modalities effectively decouples complex scene dynamics, achieving strong detection performance. 
\textcolor{black}{More broadly, temporal modeling and spatio-temporal attention are widely adopted to strengthen BEV representations under dynamic scenes, echoing insights from spatial-temporal attention in video understanding~\cite{yan2019stat} and reasoning-oriented BEV perception~\cite{shao2023reasonnet}.} 
To mitigate the high computational cost of dense BEV feature, sparse-perception methods have begun to emerge~\cite{sun2024sparsedrive,zhu2024autonomous,zhang2024sparsead,linavigation}. 
SparseDrive~\cite{sun2024sparsedrive} presents a symmetric sparse-perception module that jointly learns detection, tracking, and online mapping to produce a fully sparse scene representation. 
SparseAD~\cite{zhang2024sparsead} uses a compact query set to encode the driving scene sparsely. 
Our work follows this sparse-representation paradigm with zero annotation overhead.

\section{Method}\label{3}
\begin{figure*}[tb]
  \centering
    \vspace{-.2in}
  \includegraphics[width=0.92\linewidth]{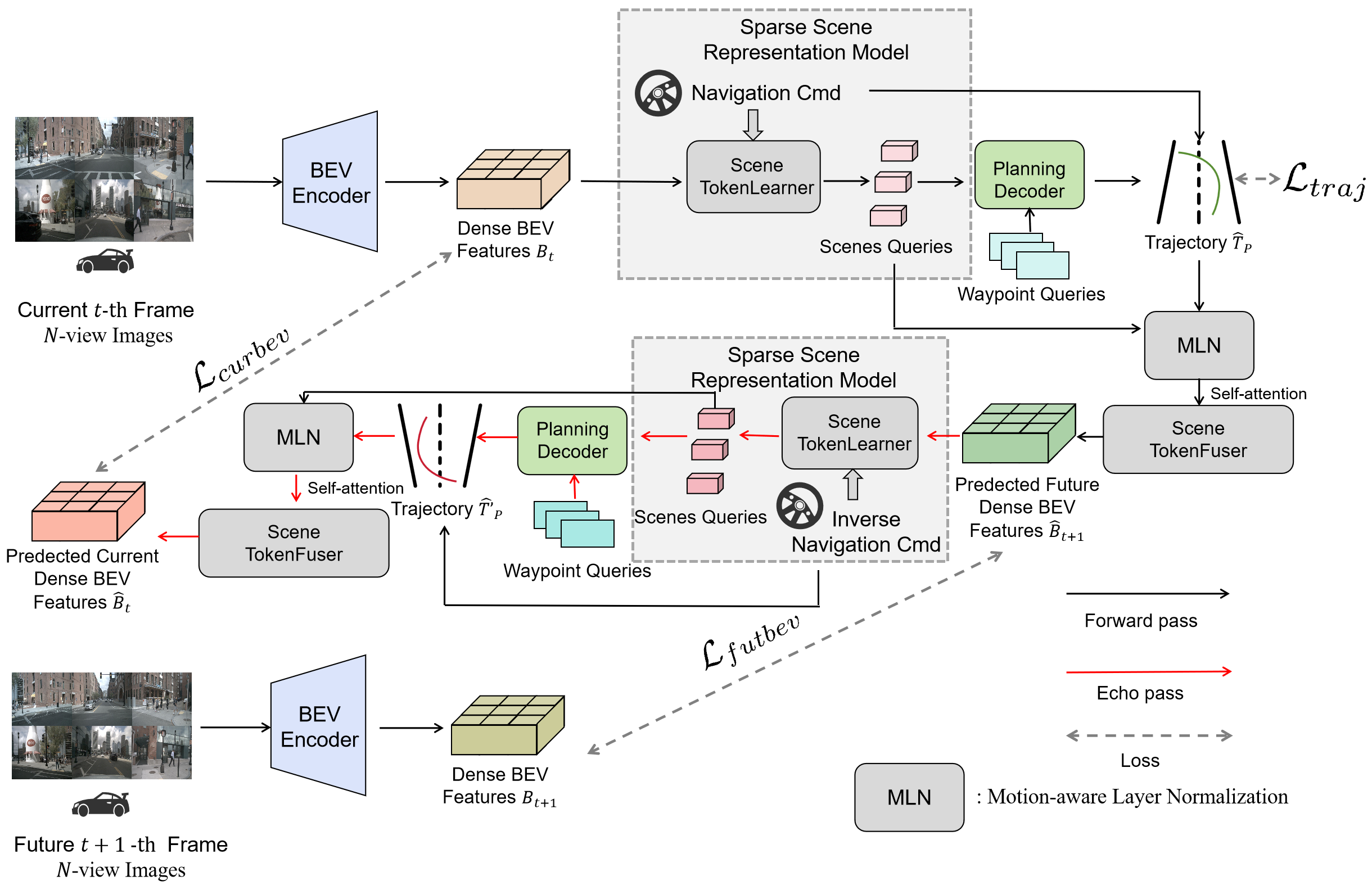}
    \vspace{-.1in}
  \caption{ \textbf{The overview of our EchoP framework.} EchoP is trained through two complementary loops. The forward loop, indicated by \uline{black arrows} in the figure, predicts future BEV features through the sparse scene representation model~\cite{linavigation} and applies self-supervision against the ground-truth BEV of future frames. The echo loop, shown in \textcolor{red}{\uline{red}}, takes those predicted future features via the Motion-aware Layer Normalization (MLN) module~\cite{wang2023exploring,linavigation} and TokenFuser~\cite{ryoo2021tokenlearner,linavigation}, reconstructs the current BEV, and self-supervises against the ground-truth BEV of the current frame.
  By validating perception in both directions along the CFC cycle, the model cross-checks its understanding of the surrounding scene and thus produces reliable trajectory plans. 
  \textbf{Notably, modules like Scene TokenLearner, Scene TokenFuser, Planning Decoder, and MLN, share weights during training. The CFC cycle does not impact inference speed, as only a forward pass from Current  →  Future is required during testing, consistent with existing methods.}
  }
  \label{fig:2}
  \vspace{-.1in}
\end{figure*}

Given raw sensor inputs (\ie, surrounding $N$-view camera images $I^{i}, i =1,..., N $) and high-level navigation command $C_{navi}$ extracted from the dataset, our Echo planning model plans the future trajectory $T$ of the ego Vehicle (see Fig.~\ref{fig:2}). EchoP first encodes the image features with a backbone network and then uses bev queries to convert image features into BEV feature space (see Sec.~\ref{3.2}). Second, EchoP distills task-critical cues from the ego-centric BEV space via a sparse scene representation module~\cite{linavigation} and condition them on the high-level navigation command, thereby mimicking the selective attention that human drivers pay to salient scene elements (Sec.~\ref{3.2}). Finally, at the core of our approach, EchoP introduces a CFC cycle: it first predicts the ego-vehicle’s future trajectory from the navigation-conditioned scene features, then inversely infers the present ego state from this forecast, and finally reconstructs the current BEV feature map (Sec.~\ref{3.3}). Unlike prior end-to-end planners that propagate information only in the forward temporal direction, this bidirectional reasoning mechanism leverages temporal consistency within the ego sequence, leveraging the sequential context and yielding more reliable planning.

\subsection{Preliminary}\label{3.2}
\textbf{BEV features generation.}
At the $t$-th time step, the surrounding $N$-view camera images $I^{i}, \left(i =1,..., N \right)$, are transformed into a BEV representation using BEVFormer~\cite{10791908}. Specifically, images $I^{i}$ pass through an image backbone to yield image features $F_{I}^{i}, \left(i =1,..., N \right)$. Retaining the BEV representation $B_{t-1}$ from the preceding timestamp $t-1$, a group of learnable BEV queries $Q\in \mathbb{R} ^{H\times W\times K} $ simultaneously extracts temporal context from the previous BEV features $B_{t-1}$ and spatial information from the current multi-camera features $\left [ F_{I}^{i} \right ] _{t}$ via cross-attention. This results in the updated BEV representation $B_{t}\in \mathbb{R} ^{H\times W\times K}$, where $H, W$ define the spatial resolution of the BEV plane, and $K$ denotes the feature channel dimensionality.

\textbf{Sparse scene representation module.}
Given the BEV features and navigation command, the sparse scene representation module, such as SSR~\cite{linavigation}, adopts the scene token learner to extract scene queries $S\in \mathbb{R} ^{N_{s}\times K}$ from dense BEV features $B$, where $N_{s}$ is the number of scene queries. Firstly, the navigation command $C_{navi}$ is encoded into the dense BEV feature~\cite{Hu_Shen_Sun_2018}. Then, the navigation-aware BEV feature goes through the BEV TokenLearner module~\cite{ryoo2021tokenlearner} and multi-layer self-attention~\cite{Vaswani_Shazeer_Parmar_Uszkoreit_Jones_Gomez_Kaiser_Polosukhin_2017} to get the current frame sparse scene representation $S_{t}$. After obtaining the current-frame scene representation $S_{t}$ that encodes the salient BEV context, we introduce a set of waypoint queries $W_{t}\in \mathbb{R} ^{N_{t}\times N_{c}\times K}$. These queries attend to $S_{t}$ to capture the prospective motion of vehicle, where $N_{t}$ is the prediction horizon (in time steps), $N_{c}$ is the number of discrete navigation commands, and $K$ is the feature dimensionality. A lightweight multi-layer perceptron subsequently regresses the waypoint embeddings into planar coordinates, producing the predicted ego trajectory $T_{P}\in \mathbb{R} ^{N_{t} \times 2}$ over the next $N_{t}$ time steps. For supervision, the high-level navigation command $C_{navi}$ selects the mode-consistent branch $\hat{T}_{P} $, which is compared with the ground-truth trajectory $T_{GT}$ using an L1 loss:
\begin{equation}
\mathcal{L} _{traj} =\left \| \hat{T}_{P} - T_{GT} \right \| _{1}. 
\label{eq:1}
\end{equation}

\subsection{Echo planning}\label{3.3}

\textbf{Motivation.} In end-to-end autonomous driving planning, real-time scene context should strictly inform the prediction of the vehicle's trajectory. However, relying solely on forward prediction is often an ill-posed problem, as the model may generate trajectories that are statistically probable but semantically inconsistent with the current fine-grained scene details. Such feed-forward approaches are prone to hallucinations when facing complex occlusions or open-set sensor noise, lacking a mechanism to verify the causality of the generated plan.
To address this, we argue that a robust planner should enforce bi-directional consistency: a plausible trajectory should not only be inferred from the current observation but should also contain sufficient information to inversely reconstruct that observation. This reversibility serves as a hard geometric and semantic constraint to filter out inconsistent predictions that deviate from the actual scene evolution. Motivated by this need for intrinsic spatio-temporal verification, we propose EchoP, a cycle-loop self-correcting framework that leverages self-supervised training across time to substantially enhance planning fidelity without requiring external supervision. As illustrated in Fig.~\ref{fig:2}, we introduce a recurrent planner that forms a CFC BEV cycle loop, continuously validating trajectory predictions by inversely reconstructing the present BEV state. \textcolor{black}{Fig.~\ref{fig:2} visualizes the full training graph of EchoP, where the \emph{forward loop} (black arrows) and the \emph{echo loop} (red arrows) share the same core modules.
From current $N$-view images, a BEV encoder produces dense BEV features $B_t$, which are compressed into sparse scene tokens $S_t$ by the Scene TokenLearner.
Conditioned on the navigation command, the planning decoder regresses the ego trajectory and is supervised by the trajectory loss in Eq.~\ref{eq:1}.
In the forward loop, the predicted future scene tokens are fused back to dense BEV $\hat{B}_{t+1}$ (TokenFuser) and are self-supervised by the future-BEV loss in Eq.~\ref{eq2}.
In the echo loop, $\hat{B}_{t+1}$ is processed again to reconstruct the current BEV $\hat{B}_t$ and is self-supervised by the current-BEV loss in Eq.~\ref{eq3}.
The dashed lines in Fig.~\ref{fig:2} indicate these three supervision signals.}


\textbf{Forward pass (Current → Future).} The navigation-consistent trajectory prediction $\hat{T}_{P} $ and the current scene embedding $S_{t}$ are first processed by a Motion-aware Layer Normalization (MLN) module~\cite{wang2023exploring,linavigation}, producing preliminary prediction queries. A self-attention block~\cite{Vaswani_Shazeer_Parmar_Uszkoreit_Jones_Gomez_Kaiser_Polosukhin_2017} then refines these queries into the future scene representation $\hat{S}_{t+1}$. Because the regions of interest in autonomous-driving scenes shift over time, we provide an additional source of supervision. The predicted future scene embedding $\hat{S}_{t+1}$ is fed to TokenFuser~\cite{ryoo2021tokenlearner,linavigation} to hallucinate a dense BEV feature map $\hat{B}_{t+1}$. We then enforce an L2 penalty between $\hat{B}_{t+1}$ and $B_{t}$ the ground-truth BEV, yielding the future-BEV reconstruction loss $\mathcal{L} _{futbev}$:
\begin{equation}
\mathcal{L} _{futbev} =\left \| \hat{B}_{t+1} -B_{t+1} \right \| _{2}. 
\label{eq2}
\end{equation}

\textbf{Echo pass (Future → Current).} Our goal in this stage is to start from the predicted dense BEV at $T+1$, $\hat{B}_{t+1}$, and run the pipeline in reverse to reconstruct the current BEV $\hat{B}_{t}$ for self-supervision. The Echo pass comprises four steps. Firstly, to mimic backward driving, we derive a reversed navigation command ${C}'_{navi}$ from the original $C_{navi}$ and feed $\left [ \hat{B}_{t+1},{C}'_{navi} \right ] $ into the same sparse-scene-representation module. The reversed navigation command ${C}'_{navi}$ is derived by inverting the original discrete command $C_{navi}$: specifically, the "left" and "right" directions are swapped, while "go straight" remains unchanged. Secondly, the scene token learner~\cite{linavigation}, equipped with an attention module, extracts predicted future scene queries $\hat{S}'_{t+1}$ from the input BEV features $\hat{B}_{t+1}$. Thirdly, a new set of waypoint queries ${W}'_{t+1}$ attends to $\hat{S}'_{t+1}$, generating a reversed trajectory proposal $\hat{T}'_{p}$, after which the inverse command ${C}'_{navi}$ selects the mode-consistent branch of $\hat{T}'_{p}$. Finally, analogous to the forward pass, $\hat{T}'_{p}$ and $\hat{S}'_{t+1}$ are processed by an MLN block followed by stacked self-attention layers, producing the current-frame scene queries $\hat{S}'_{t}$. TokenFuser~\cite{ryoo2021tokenlearner,linavigation} then reconstructs the dense BEV map $\hat{B}_{t}$ from $\hat{S}'_{t}$.
The discrepancy between the reconstructed and ground-truth BEV features derives the current BEV reconstruction loss as follow:
\begin{equation}
\mathcal{L} _{curbev} =\left \| \hat{B}_{t} -B_{t} \right \| _{2}. 
\label{eq3}
\end{equation}

\textbf{Total loss.} In summary, EchoP is trained end‑to‑end with a composite objective comprising the trajectory loss $\mathcal{L}_{traj}$, the future‑BEV reconstruction loss $\mathcal{L} _{futbev}$, and the current‑BEV reconstruction loss $\mathcal{L}_{curbev}$, combined as: $\mathcal{L} _{total} = \mathcal{L} _{traj} + \lambda_{futbev} \mathcal{L} _{futbev} + \lambda_{curbev} \mathcal{L} _{curbev}.$
\textcolor{black}{\subsection{Method Roadmap and Practical Interpretation}}
\label{sec:roadmap}
\textcolor{black}{To make the construction of EchoP easier to follow, we summarize the method as a two-pass training procedure over a
\textbf{Current$\rightarrow$Future$\rightarrow$Current (CFC)} cycle.
\textbf{Forward pass (Current$\rightarrow$Future).}
Given the current BEV feature $B_t$ and the navigation command $C_{\text{navi}}$, EchoP predicts a navigation-consistent
trajectory $\hat{T}_P$ and a future scene representation $\hat{S}_{t+1}$, which is further decoded into a dense future BEV
$\hat{B}_{t+1}$. This stage is supervised by the future-BEV reconstruction loss $L_{\text{futbev}}$ (Eq.~\ref{eq2}).
\textbf{Echo pass (Future$\rightarrow$Current).}
Starting from the predicted future BEV $\hat{B}_{t+1}$, EchoP runs a mirrored pipeline (with shared weights) to reconstruct
the current BEV $\hat{B}_t$, supervised by the current-BEV reconstruction loss $L_{\text{curbev}}$ (Eq.~\ref{eq3}).
This cycle provides an intrinsic spatio-temporal consistency check: a plausible plan should be reversible enough to
reconstruct the observation it is conditioned on.
\textbf{Deployment note.}
The echo pass is used \emph{only during training}. At inference time, EchoP performs a single forward pass
(Current$\rightarrow$Future) to output $\hat{T}_P$, incurring no additional runtime overhead compared with one-shot planners.}

\begin{table*}[t]
\centering
\vspace{-.2in}
\caption{Equation-to-module mapping and how each term relates to evaluation metrics.}
\label{tab:eq_mapping}
\begin{tabular}{lccc}
\toprule
Component & Output & Supervision & Main effect on metrics \\
\midrule
Trajectory regression $L_{\text{traj}}$ (Eq.~(1)) &
$\hat{T}_P$ &
$T_{\text{GT}}$ &
Reduces L2 displacement error \\
Future BEV reconstruction $L_{\text{futbev}}$ (Eq.~(2)) &
$\hat{B}_{t+1}$ &
$B_{t+1}$ &
Improves future scene grounding \\
Current BEV reconstruction $L_{\text{curbev}}$ (Eq.~(3)) &
$\hat{B}_{t}$ &
$B_{t}$ &
Enforces temporal consistency; reduces collisions \\
\bottomrule
\end{tabular}
\end{table*}

\section{Experiment}\label{4}
In this section, we provide a comprehensive evaluation of the proposed EchoP framework, organized into three main parts: experimental setup, comparative analysis, and in-depth discussion. We begin by introducing the standard open-loop benchmark nuScenes~\cite{caesar2020nuscenes} and the closed-loop benchmark Bench2Drive~\cite{jia2024bench}, followed by the definitions of key evaluation metrics (e.g., L2 error, collision rate, and driving score) and the implementation specifics. The extensive comparison with state-of-the-art end-to-end planners is presented in Section~\ref{sec:comparison}. Finally, Sections~\ref{sec:ablation} and~\ref{sec:qualitative} discuss the ablation studies validating the spatio-temporal consistency of the CFC cycle and provide qualitative visualizations of planning behaviors, respectively.

\textbf{Dataset.}
We conduct extensive experiments on the widely adopted nuScenes dataset~\cite{caesar2020nuscenes} to assess the open-loop planning capability of our EchoP framework. nuScenes comprises 1,000 driving logs, partitioned into 700, 150, and 150 scenes for training, validation, and testing, respectively. Each scene provides 20 seconds (s) of synchronized RGB and LiDAR data captured at 12 Hz, supplemented by key-frame annotations at 2 Hz. The sensor suite includes six surround-view cameras covering a full 360° field of view and a 32-beam LiDAR. The dataset supplies dense semantic maps alongside 1.4M 3D bounding boxes spanning 23 object classes, providing a rich testbed for perception and prediction. \textcolor{black}{
\noindent\textbf{Real-world factors in nuScenes.}
We emphasize that our main open-loop benchmark is \emph{not} confined to simulation: nuScenes is collected from real driving logs with multi-view RGB streams, and contains diverse real-world conditions (e.g., rainy scenes and low-light nighttime driving). Therefore, our evaluation already covers practical visual challenges beyond idealized settings, as also illustrated by the qualitative cases in Fig.~\ref{fig:3}(b, d).
}

To rigorously evaluate the planning robustness in an interactive environment, we further employ Bench2Drive~\cite{jia2024bench}, a state-of-the-art closed-loop benchmark built upon the CARLA simulator. Unlike open-loop evaluation, Bench2Drive requires the agent to actively control the vehicle across 220 official routes, navigating through dynamic traffic and complex topologies under diverse weather conditions. This benchmark features a multi-ability evaluation protocol that dissects driving performance into distinct competencies, such as lane changing, intersection traversal, and yielding, thereby providing a granular and comprehensive assessment of the planner's generalization capability and safety in long-horizon execution. \textcolor{black}{
We stress that Bench2Drive complements open-loop evaluation by enforcing \emph{closed-loop} execution, where the agent must repeatedly re-plan under dynamic traffic and long-horizon interactions, thereby partially bridging the gap between offline metrics and real-world operational constraints.
}

\begin{table*}[t]
  \centering
  \footnotesize
  \vspace{-.2in}
  \setlength{\tabcolsep}{6.0pt}
  \caption{Open-Loop Planning on nuScenes. The \textbf{top block} with \textsuperscript{o} means the lidar-based methods. The  \textbf{middle block} follows the UniAD protocol (final/max aggregation). The  \textbf{lower block} with \textsuperscript{\ddag} follows VAD protocol (average over all predicted frames). $^\star$: The backbone is ResNet-101~\cite{he2016deep}, while other methods without $^\star$ adopt ResNet-50 or similar variants. \textsuperscript{\S}: The result that we re-implement SSR with official weights~\cite{linavigation}. The $\downarrow$ indicates that lower is better.}
  \label{tab:1}
  \renewcommand{\arraystretch}{1.2}
  \footnotesize
  \begin{tabular}{l l cccc cccc}
    \toprule
    \multirow{2}{*}{Method} & \multirow{2}{*}{Auxiliary Task} 
      & \multicolumn{4}{c}{\textbf{L2} (m) $\downarrow$}
      & \multicolumn{4}{c}{\textbf{Collision Rate} (\%) $\downarrow$} \\
    \cmidrule(lr){3-6} \cmidrule(lr){7-10}
    & & 1s & 2s & 3s & Avg.
      & 1s & 2s & 3s & Avg. \\
    \midrule
    NMP\textsuperscript{o}~\cite{zeng2019end}     & Det \& Motion                & 0.53 & 1.25 & 2.67 & 1.48 & 0.04 & 0.12 & 0.87 & 0.34 \\
    FF\textsuperscript{o}~\cite{hu2021safe}        & FreeSpace                    & 0.55 & 1.20 & 2.54 & 1.43 & 0.06 & 0.17 & 1.07 & 0.43 \\
    EO\textsuperscript{o}~\cite{khurana2022differentiable}   & FreeSpace                    & 0.67 & 1.36 & 2.78 & 1.60 & 0.04 & 0.09 & 0.88 & 0.33 \\ 
    \midrule
    ST\textendash P3~\cite{hu2022st}             & Det \& Map \& Depth          & 1.72 & 3.26 & 4.86 & 3.28 & 0.44 & 1.08 & 3.01 & 1.51 \\
    UniAD\textsuperscript{$\star$}~\cite{hu2023planning} & {\scriptsize Det\&Track\&Map\&Motion\&Occ}   & 0.48 & 0.96 & 1.65 & 1.03 & 0.05 & 0.17 & 0.71 & 0.31 \\
    OccNeXt\textsuperscript{$\star$}~\cite{tong2023scene}  
                                               & Det \& Map \& Occ            & 1.29 & 2.13 & 2.99 & 2.14 & 0.21 & 0.59 & 1.37 & 0.72 \\
    VAD\textendash Base~\cite{jiang2023vad}       & Det \& Map \& Motion         & 0.54 & 1.15 & 1.98 & 1.22 & 0.04 & 0.39 & 1.17 & 0.53 \\
    PARA\textendash Drive~\cite{weng2024drive}      & {\scriptsize Det\&Track\&Map\&Motion\&Occ}     & 0.40 & 0.77 & 1.31 & 0.83 & 0.07 & 0.25 & 0.60 & 0.30 \\
    GenAD~\cite{zheng2024genad}                    & Det \& Map \& Motion         & 0.36 & 0.83 & 1.55 & 0.91 & 0.06 & 0.23 & 1.00 & 0.43 \\
    SparseDrive~\cite{sun2024sparsedrive}         & {\scriptsize Det\&Track\&Map\&Motion}         & 0.44 & 0.92 & 1.69 & 1.01 & 0.07 & 0.19 & 0.71 & 0.32 \\
    UAD\textendash Tiny~\cite{guo2024end}         & Det                          & 0.47 & 0.99 & 1.71 & 1.06 & 0.08 & 0.39 & 0.90 & 0.46 \\
    UAD\textsuperscript{$\star$}~\cite{guo2024end} & Det                         & 0.39 & 0.81 & 1.50 & 0.90 & \textbf{0.01} & 0.12 & 0.43 & 0.19 \\
      
    SSR\textsuperscript{\S}~\cite{linavigation}  & None                        & 0.25 &0.64 & 1.33 & 0.74
      & 0.16 & 0.21 & 0.51 & 0.29 \\
    MomAD~\cite{song2025don}  & None                        & 0.43 &0.88 & 1.62 & 0.98
      & 0.06 & 0.16 & 0.68 & 0.30 \\
    \textbf{EchoP (Ours) } & None & \textbf{0.23} & \textbf{0.60} & \textbf{1.27} & \textbf{0.70} (\textcolor{red}{-0.04})
          & 0.02 & \textbf{0.12} & \textbf{0.39} & \textbf{0.17} (\textcolor{red}{-0.12}) \\        
    \midrule
    \rowcolor{Gray}
    ST\textendash P3\textsuperscript{\ddag}~\cite{hu2022st}      & Det \& Map \& Depth          & 1.33 & 2.11 & 2.90 & 2.11 & 0.23 & 0.62 & 1.27 & 0.71 \\
    \rowcolor{Gray}
    UniAD\textsuperscript{$\star\ddag$}~\cite{hu2023planning}        & {\scriptsize Det\&Track\&Map\&Motion\&Occ}     & 0.44 & 0.67 & 0.96 & 0.69 & 0.04 & 0.08 & 0.23 & 0.12 \\
    \rowcolor{Gray}
    VAD\textendash Tiny\textsuperscript{\ddag}~\cite{jiang2023vad}   & Det \& Map \& Motion         & 0.46 & 0.76 & 1.12 & 0.78 & 0.21 & 0.35 & 0.58 & 0.38 \\
    \rowcolor{Gray}
    VAD\textendash Base\textsuperscript{\ddag}~\cite{jiang2023vad}  & Det \& Map \& Motion         & 0.41 & 0.70 & 1.05 & 0.72 & 0.07 & 0.17 & 0.41 & 0.22 \\
    \rowcolor{Gray}
    HE\textendash Drive\textsuperscript{\ddag}~\cite{wang2024he} & {\scriptsize Det\&Track\&Map}     & 0.31 & 0.58 & 0.93 & 0.60 & 0.01 & 0.05 & 0.16 & 0.07 \\
    \rowcolor{Gray}
    BEV\textendash Planner\textsuperscript{\ddag}~\cite{li2024ego} & None                        & 0.28 & 0.42 & 0.68 & 0.46 & 0.04 & 0.37 & 1.07 & 0.49 \\
    \rowcolor{Gray}
    PARA\textendash Drive\textsuperscript{\ddag}~\cite{weng2024drive} & {\scriptsize Det\&Track\&Map\&Motion\&Occ}     & 0.25 & 0.46 & 0.74 & 0.48 & 0.14 & 0.23 & 0.39 & 0.25 \\
    \rowcolor{Gray}
    LAW\textsuperscript{\ddag}~\cite{li2024enhancing}                 & None                         & 0.26 & 0.57 & 1.01 & 0.61 & 0.14 & 0.21 & 0.54 & 0.30 \\
    \rowcolor{Gray}
    GenAD\textsuperscript{\ddag}~\cite{zheng2024genad}            & Det \& Map \& Motion         & 0.28 & 0.49 & 0.78 & 0.52 & 0.08 & 0.14 & 0.34 & 0.19 \\
    \rowcolor{Gray}
    SparseDrive\textsuperscript{\ddag}~\cite{sun2024sparsedrive}         & {\scriptsize Det\&Track\&Map\&Motion}         & 0.29 & 0.58 & 0.96 & 0.61 & 0.01 & 0.05 & 0.18 & 0.08 \\
    \rowcolor{Gray}
    UAD\textsuperscript{$\star\ddag$}~\cite{guo2024end}         & Det                          & 0.28 & 0.41 & 0.65 & 0.45 & 0.01 & \textbf{0.03} & 0.14 & \textbf{0.06} \\
    \rowcolor{Gray}
    DiffDrive \textsuperscript{\ddag}~\cite{Liao2024DiffusionDriveTD} & {\scriptsize Det \&Map}     & 0.27 & 0.54 & 0.90 & 0.57 & 0.03 & 0.05 & 0.16 & 0.08 \\
    \rowcolor{Gray}
  
    \rowcolor{Gray}
    SSR\textsuperscript{\S}\textsuperscript{\ddag}~\cite{linavigation}   & None                         & 0.19 & 0.36 & 0.62 & 0.39
      & 0.10 & 0.13 & 0.22 & 0.15 \\
    \rowcolor{Gray}
    MomAD~\textsuperscript{\ddag}~\cite{song2025don}  & None                        & 0.31 &0.57 & 0.91 & 0.60
      & 0.01 & 0.05 & 0.22 & 0.09 \\  
    \rowcolor{Gray}
    BridgeAD~\textsuperscript{\ddag}~\cite{zhang2025bridging}  & None                        & 0.28 &0.55 & 0.92 & 0.58
      & 0.00 & 0.04 & 0.20 & 0.08 \\  
    \rowcolor{Gray}
    \textcolor{black}{FocalAD~\textsuperscript{\ddag}~\cite{focalad2025}}  & \textcolor{black}{Det \& Map \& Motion}  & \textcolor{black}{0.27} &\textcolor{black}{0.57} & \textcolor{black}{0.96} & \textcolor{black}{0.60}
      & \textcolor{black}{\textbf{0.00}} & \textcolor{black}{0.04} & \textcolor{black}{0.24} & \textcolor{black}{0.09} \\  
    \rowcolor{Gray}
    \textbf{EchoP\textsuperscript{\ddag} (Ours) } & None & \textbf{0.17} & \textbf{0.33} & \textbf{0.58} & \textbf{0.36} (\textcolor{red}{-0.03})
    & 0.03 & 0.05 & \textbf{0.14} &0.07 (\textcolor{red}{-0.08}) \\                     
    \bottomrule
  \end{tabular}
  \vspace{-.1in}
\end{table*}

\textbf{Evaluation metrics.}
Following prior work, we evaluate open-loop performance on nuScenes using two primary metrics: L2 displacement error (the Euclidean distance between the planned and ground-truth trajectories) and Collision Rate (the proportion of planned trajectories that collide with any traffic participant). Unless specified otherwise, the model is fed with 2 seconds of history corresponding to five frames, and planning quality is reported for 1s, 2s, and 3s horizons. Because VAD~\cite{jiang2023vad} and UniAD~\cite{hu2023planning} adopt different aggregation rules, we report results under both protocols: VAD averages over all predicted frames, whereas UniAD uses the final value (maximum collision). Table~\ref{tab:1} presents the outcomes under both protocols.

For Bench2Drive closed-loop evaluation, we report: Success Rate (SR), the percentage of failure-free route completions; Driving Score (DS), the primary metric weighting completion by infraction penalties; and Efficiency, which evaluates trajectory smoothness and travel time relative to the standard.

\textbf{Implementation details.}
EchoP is trained for 12 epochs on 4 RTX-A6000 GPUs with a batch size of 1 per GPU. The image encoder is a ResNet-50~\cite{he2016deep} fed with 640 × 360 images, yielding a 100 × 100 BEV grid and 16 × 256 sparse tokens. The navigation-command set remains 3, matching prior work. For training, we deploy AdamW~\cite{loshchilov2017decoupled} with a learning rate of $5\times10^{-5}$, and all other hyperparameters follow SSR~\cite{linavigation}. Unless stated, loss weights for future-BEV, and current-BEV are $\lambda_{futbev}=0.5$, and $\lambda_{curbev}=0.1$ by default.
\textcolor{black}{\noindent\textbf{Runtime and throughput.}
We report end-to-end inference speed on an NVIDIA A100-PCIE-40GB using the standard speed-test protocol (world size $=1$, batch size $=1$).
Our implementation achieves $46.27$ ms/iter, corresponding to $21.61$ FPS.
The overhead is limited because the CFC branch reuses the BEV latent features and shares backbone computation, while the inverse reconstruction is implemented as a lightweight head on compact BEV representations rather than an additional full perception stack.}

\textcolor{black}{
\textbf{Scalability and feasibility in large SDV networks.}
EchoP is designed as an on-board planner that consumes only the ego vehicle’s local multi-view observations and a high-level navigation command, and outputs short-horizon waypoints.
From a systems perspective, this implies favorable scalability in large dynamic SDV fleets: the computational cost is \emph{per-vehicle} and does not require centralized coordination for inference.
Moreover, our planning head operates on compact scene tokens (e.g., $16 \times 256$ by default) distilled from the BEV representation, which helps keep the planning computation lightweight.
As a result, EchoP can operate in real time under the measured single-GPU throughput, while retaining the benefit of the CFC self-supervision during training.
}

\begin{table}[t]
  \centering
  \vspace{-.2in}
  \setlength{\tabcolsep}{1.0pt}
  \caption{Open-loop and Closed-Loop Planning on Bench2Drive. The $\uparrow$ indicates that higher is better. The $\downarrow$ indicates that lower is better. \textsuperscript{\S}: The result that we retrain the SSR~\cite{linavigation}.}
  \label{tab:5}
  \vspace{-.05in}
  \footnotesize
  \begin{tabular}{lcccc}
    \toprule
    & \multicolumn{1}{c}{Open-loop} & \multicolumn{3}{c}{Closed-loop} \\
    \cmidrule(lr){2-2}\cmidrule(lr){3-5}
    Method & Avg.\ L2 $\downarrow$ & Driving score $\uparrow$ & Success Rate (\%) $\uparrow$ & Efficiency $\uparrow$ \\
    \midrule
    AD-MLP~\cite{zhai2023rethinkingopenloopevaluationendtoend} & 3.64 & 18.05 & 0.00  & 48.45  \\
    UniAD-Tiny~\cite{hu2023planning}                           & 0.80 & 40.73 & 13.18 & 123.92 \\
    UniAD-Base~\cite{hu2023planning}                           & \textbf{0.73} & 45.81 & 16.36 & 129.21 \\
    VAD~\cite{jiang2023vad}                                    & 0.91 & 42.35 & 15.00 & 157.94 \\
    GenAD~\cite{zheng2024genad}                                    & - & 44.81 & 15.90 & -   \\
    SSR\textsuperscript{\S}~\cite{linavigation}                                    & 0.90 & 32.34 & 11.85 & 157.62 \\
    MomAD~\cite{song2025don}                              & 0.85 & 45.35 & 17.44 & 162.09 \\
    \textcolor{black}{FocalAD~\cite{focalad2025}}                        & \textcolor{black}{0.85} & \textcolor{black}{45.77} & \textcolor{black}{17.30} & \textcolor{black}{174.01} \\
    \textcolor{black}{DIVER~\cite{diver2025}}                        & \textcolor{black}{1.05} & \textcolor{black}{49.21} & \textcolor{black}{21.56} & \textcolor{black}{177.00} \\
    \rowcolor{Gray}
    EchoP (ours)                                                 & 0.79 & \textbf{50.35} & \textbf{26.54} & \textbf{185.18} \\
    \bottomrule
  \end{tabular}
  \vspace{-.15in}
\end{table}

\begin{table}[t] 
  \centering
  \vspace{-.2in}
    \centering
    \setlength{\tabcolsep}{2pt}
    \scriptsize
    \vspace{-.05in}
    \renewcommand{\arraystretch}{1.2}
    \caption{Ablation for our CFC cycle. The upper block follows the UniAD protocol (final/max aggregation). The lower block with \textsuperscript{\ddag} follows VAD protocol (average over all predicted frames).\textsuperscript{\S}: The result that we retrain the SSR~\cite{linavigation} on 4 GPUs. The $\downarrow$ indicates that lower is better.}
    \begin{tabular}{l c cccc cccc}
      \toprule
      \multirow{2}{*}{Method} & \multirow{2}{*}{CFC cycle} 
        & \multicolumn{4}{c}{\textbf{L2} (m) $\downarrow$}
        & \multicolumn{4}{c}{\textbf{Collision Rate} (\%) $\downarrow$} \\
      \cmidrule(lr){3-6} \cmidrule(lr){7-10}
      & & 1s & 2s & 3s & Avg.
        & 1s & 2s & 3s & Avg. \\
      \midrule
      SSR\textsuperscript{\S}~\cite{linavigation}   &                       & 0.25 & 0.64 & 1.33 & 0.74
        & 0.16 & 0.21 & 0.51 & 0.29 \\
      \textbf{EchoP (Ours)} & \checkmark & \textbf{0.23} & \textbf{0.60} & \textbf{1.27} & \textbf{0.70}
            & \textbf{0.02} & \textbf{0.12} & \textbf{0.39} & \textbf{0.17} \\       
      \midrule
      \rowcolor{Gray}
      SSR\textsuperscript{\S}\textsuperscript{\ddag}~\cite{linavigation}  &                       & 0.19 & 0.36 & 0.62 & 0.39
        & 0.10 & 0.13 & 0.22 & 0.15 \\
      \rowcolor{Gray}
      \textbf{EchoP\textsuperscript{\ddag} (Ours)} & \checkmark & \textbf{0.17} & \textbf{0.33} & \textbf{0.58} & \textbf{0.36}
      & \textbf{0.03} & \textbf{0.05} & \textbf{0.14} & \textbf{0.07} \\                     
      \bottomrule
    \end{tabular}
    \label{tab:2}
    \vspace{-.15in}
\end{table}

\begin{table}[t] 
  \centering
  \vspace{-.1in}
    \centering
    \setlength{\tabcolsep}{4.0pt}
    \scriptsize
    \renewcommand{\arraystretch}{1.2}
        \caption{Ablation for the CFC cycle effectiveness in different scene representation numbers. The block all follows the UniAD protocol (final/max aggregation). The row with a gray background is the re-implementation of our baseline. The $\downarrow$ indicates that lower is better.}
    \begin{tabular}{c c cccc cccc}
      \toprule
      \multirow{2}{*}{$N_{s}$} & \multirow{2}{*}{CFC cycle}
        & \multicolumn{4}{c}{\textbf{L2} (m) $\downarrow$}
        & \multicolumn{4}{c}{\textbf{Collision Rate} (\%) $\downarrow$} \\
      \cmidrule(lr){3-6} \cmidrule(lr){7-10}
      & & 1s & 2s & 3s & Avg.
       & 1s & 2s & 3s & Avg. \\
      \midrule
      \rowcolor{Gray}
    8 &   & 0.27 & 0.69 & 1.43 & 0.79
    & 0.20 & 0.35 & 0.68 & 0.41 \\
      8 & \checkmark & \textbf{0.21} & \textbf{0.57} & \textbf{1.22} & \textbf{0.67}
        & \textbf{0.00} & \textbf{0.06} & \textbf{0.45} & \textbf{0.17} \\       
      \midrule
      \rowcolor{Gray}
      16  &   & 0.25 & 0.64 & 1.33 & 0.74
        & 0.16 & 0.21 & 0.51 & 0.29 \\
      16 & \checkmark & \textbf{0.23} & \textbf{0.60} & \textbf{1.27} & \textbf{0.70}
            & \textbf{0.02} & \textbf{0.12} & \textbf{0.39} & \textbf{0.17} \\       
      \bottomrule
    \end{tabular}
    \label{tab:3}
    \vspace{-.15in}
\end{table}
\begin{table}[t] 
  \centering   
  \vspace{-.2in}
  \caption{Temporal consistency on the test set. We measure the $\mathcal{L} _{curbev} =\left \| \hat{B}_{t} -B_{t} \right \| _{2}.$ Lower is better. We could observe that our method ensure the bidirectional consistency on the unseen test set.}
  \label{tab:6}
  \footnotesize
  \begin{tabular}{lc}
    \toprule
    Method & $\mathcal{L} _{curbev} \downarrow$ \\
    \midrule
    Baseline & 0.2442 \\
    EchoP (CFC cycle) & \textbf{0.0962} \\
    \bottomrule
  \end{tabular}
  \vspace{-.1in}
\end{table}

\begin{table}[t] 
  \centering    
  \setlength{\tabcolsep}{3pt} 
  \caption{Ablation for the loss weight. Varying $\mathcal{L}_{curbev}$ and $\mathcal{L}_{futbev}$.}
  \label{tab:4}
  \footnotesize
  \renewcommand{\arraystretch}{1.2}
  \begin{tabular}{c c cccc cccc}
    \toprule
    \multirow{2}{*}{$\lambda_{curbev}$} & \multirow{2}{*}{$\lambda_{futbev}$}
    & \multicolumn{4}{c}{\textbf{L2} (m) $\downarrow$}
    & \multicolumn{4}{c}{\textbf{Collision Rate} (\%) $\downarrow$} \\
    \cmidrule(lr){3-6}\cmidrule(lr){7-10}
    & & 1s & 2s & 3s & Avg. & 1s & 2s & 3s & Avg. \\
    \midrule
    0.1 & 0.5 & \textbf{0.23} & \textbf{0.60} & \textbf{1.27} & \textbf{0.70} & 0.02 & 0.12 & 0.39 & 0.17 \\
    0.5 & 0.5 & 0.24 & 0.63 & 1.34 & 0.73 & \textbf{0.00} & \textbf{0.10} & \textbf{0.35} & \textbf{0.15} \\
    0.1 & 0.8 & 0.24 & 0.63 & 1.31 & 0.72 & 0.04 & 0.14 & 0.45 & 0.21 \\
    \midrule
    \rowcolor{Gray}
    0.1 & 0.5 & \textbf{0.17} & \textbf{0.33} & \textbf{0.58} & \textbf{0.36} & \textbf{0.03} & \textbf{0.05} & \textbf{0.14} & \textbf{0.07} \\
    \rowcolor{Gray}
    0.5 & 0.5 & 0.19 & 0.36 & 0.64 & 0.39 & 0.08 & 0.06 & 0.13 & 0.09 \\
    \rowcolor{Gray}
    0.1 & 0.8 & 0.18 & 0.35 & 0.61 & 0.38 & 0.00 & 0.05 & 0.21 & 0.09 \\
    \bottomrule
  \end{tabular}
  \vspace{-.15in}
\end{table}

\subsection{Comparison with the State-of-the-art Methods}\label{sec:comparison}
\textbf{Open-Loop Evaluation.}
To assess the effectiveness of EchoP, we train and evaluate the model on the widely used nuScenes benchmark. As summarised in Table~\ref{tab:1}, EchoP achieves the best results among contemporary end-to-end planners on both the L2 error and the collision rate, attaining an average $L2_{MAX}$ error of only 0.70 m.
Compared to strong one-shot planners, our approach delivers consistent benefits. Against UniAD the average $L2_{MAX}$ error drops by 0.33 m (32\% relatively) and the average $CR_{MAX}$ falls by 0.14\% (45\% relatively). Compared with VAD-Base, the gains are even larger, with reductions of 0.52 m in average $L2_{MAX}$ (43\% relatively) and 0.36\% in average $CR_{MAX}$ (68\% relatively). In addition, we compare our approach to the SSR baseline. Due to recent updates in the SSR codebase, our reproduced results on a 4-GPU setup differ from those reported in the original paper, a point also acknowledged by the authors on GitHub, as their modifications affected collision rate performance. For a fair comparison, we benchmark EchoP against our reproduced SSR results and observe substantial improvements in both L2 error and collision rate. EchoP still lowers the average $L2_{MAX}$ by 0.04 m and the average $CR_{MAX}$ by 0.12\%, while also improving the $L2_{AVG}$ and $CR_{AVG}$ metrics by 0.03 m and 0.08\% respectively. The consistent improvements across all indicators confirm that the Current → Future → Current cycle provides a reliable mechanism for accurate and safe trajectory planning.

\textbf{Closed-Loop Evaluation.}
To evaluate the effectiveness of our approach under closed-loop control, we follow the Bench2Drive benchmark~\cite{jia2024bench}. As shown in Table~\ref{tab:5}, EchoP markedly surpasses the SSR baseline~\cite{linavigation} in Driving Score. Relative to widely used methods, EchoP attains +4.54 and +8.00 points over UniAD~\cite{hu2023planning} and VAD~\cite{jiang2023vad} , respectively. With the proposed CFC cycle enabled, EchoP further lifts the closed-loop Success Rate by +10.18\% compared to UniAD~\cite{hu2023planning}. Moreover, because EchoP does not rely on auxiliary tasks, it achieves superior runtime efficiency while maintaining high success rates.

\subsection{Ablation Studies and Further Discussion}\label{sec:ablation}
\textcolor{black}{\textbf{Connecting methodology to evaluation (open-loop and closed-loop).}
The three-term objective in total loss is designed to improve not only trajectory accuracy but also scene-consistent planning.
$L_{\text{traj}}$ directly optimizes waypoint regression (L2 displacement), while $L_{\text{futbev}}$ encourages future scene grounding.
Crucially, the CFC echo term $L_{\text{curbev}}$ regularizes temporal consistency by requiring that the predicted future state can
reconstruct the current observation, which reduces physically implausible plans and thus lowers collision rates.
In Tab.~\ref{tab:eq_mapping}, this mechanism is reflected by improved temporal consistency, open-loop L2/collision reductions,
and stronger closed-loop robustness on Bench2Drive.}

\textbf{Effect of the CFC cycle.} Table~\ref{tab:2} details the ablation of the CFC BEV cycle loop. Incorporating the CFC cycle into the one-shot planner reduces the average $L2_{MAX}$ error by 0.04 m and $CR_{MAX}$ by 0.12\%, while simultaneously lowering the average $L2_{AVG}$ and $CR_{AVG}$ by 0.03 m and 0.08\%, respectively. These results underscore the significant efficacy of bidirectional validation via inverse reconstruction, which markedly improves the understanding of dynamic scenes. Consequently, EchoP facilitates safer trajectory generation and delivers substantial planning enhancements. To further quantify the temporal consistency, we compute the Mean Squared Error (MSE) between the Current $\to$ Future $\to$ Current predicted BEV and the ground-truth Current BEV on the test set. As shown in Table~\ref{tab:6}, we benchmark the baseline against our EchoP equipped with the CFC mechanism.

\begin{figure*}[t]
  \centering
   \vspace{-.2in}
  \includegraphics[width=0.95\linewidth]{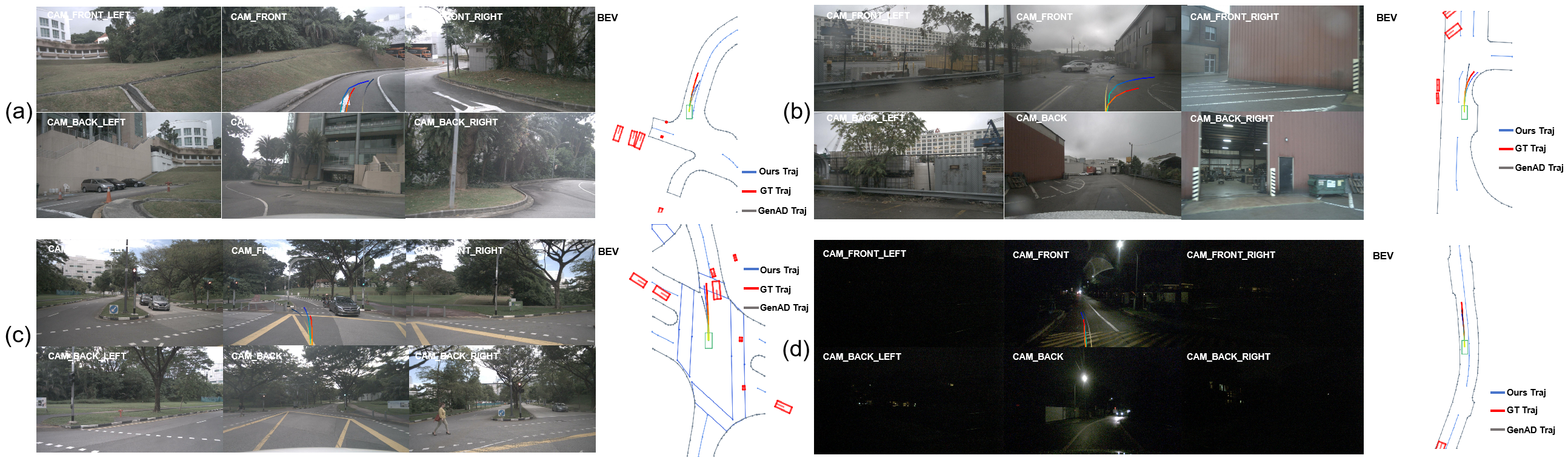}
  \vspace{-.1in}
  \caption{ \textbf{Visualization of planning trajectories in different scenes.} We visualize several driving scenes, overlaying the trajectories predicted by our method alongside those of one of the one-shot planners, \ie, GenAD~\cite{zheng2024genad}, and the ground-truth trajectories. Map context is rendered directly from the dataset annotations. In each figure, the {\color{ForestGreen}{green}} icon denotes the ego vehicle, while \textcolor{red}{red} icons mark surrounding vehicles and other dynamic objects. Video is in \textbf{Suppl.} 
  }
  \label{fig:3}
\end{figure*}

\begin{figure*}[t]
  \centering
   \vspace{-.1in}
  \includegraphics[width=0.9\linewidth]{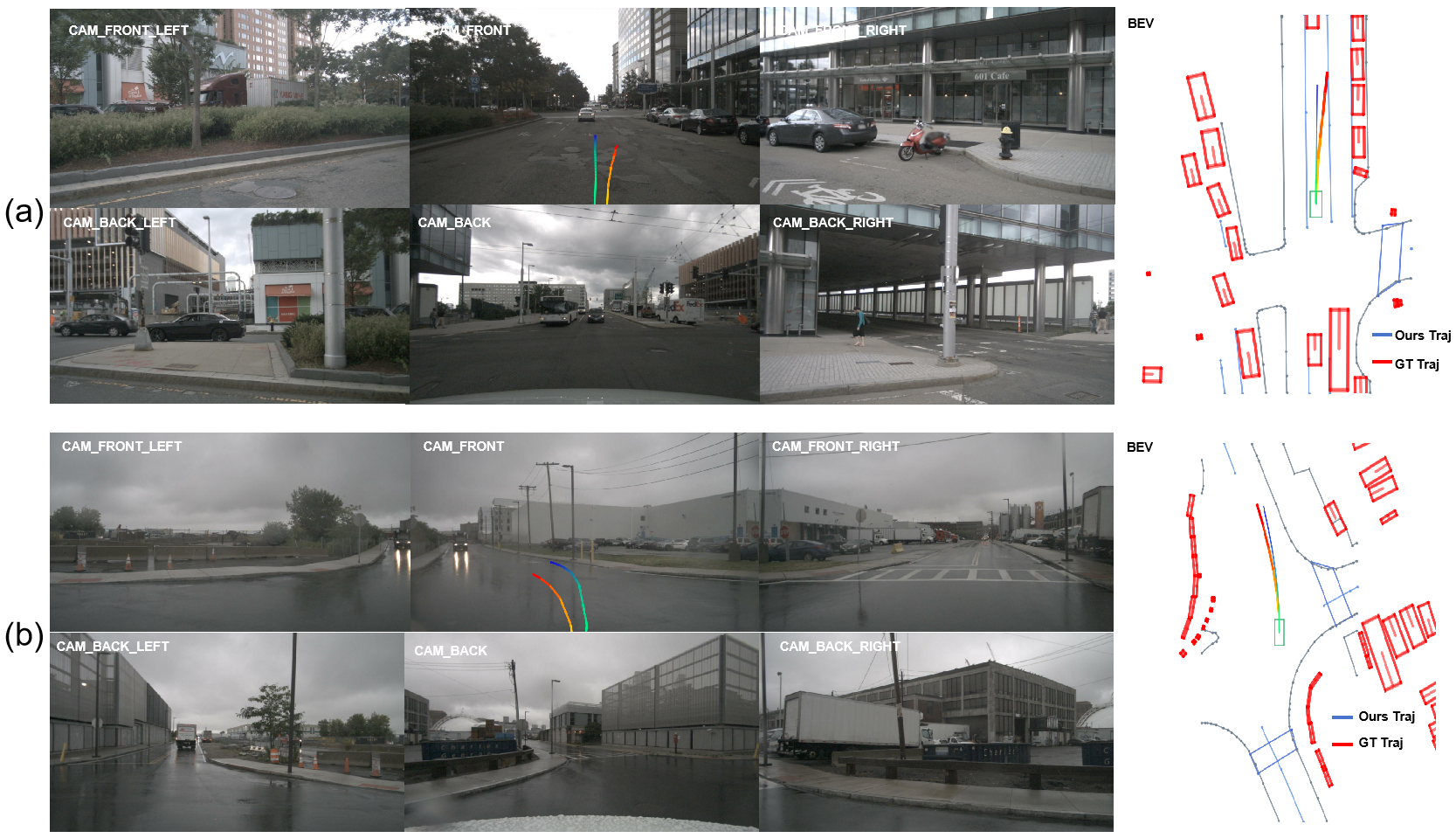}
  \vspace{-.1in}
  \caption{ \textbf{Visualization of failure cases.} We highlight two distinct planning failure cases. The first arises when the navigation cue in the ground-truth annotation is ambiguous, and the second occurs in open spaces where the predicted trajectory deviates during a turn. In each figure, the {\color{ForestGreen}{green}} icon denotes the ego vehicle, while \textcolor{red}{red} icons mark surrounding vehicles and other dynamic objects.  
  }
 \vspace{-.15in}
  \label{fig:4}
\end{figure*}

\textbf{Effect of different scene representation numbers.} In Table~\ref {tab:3}, we assess the robustness of the CFC cycle when the scene representation is compressed to different capacities. We compare two settings: 8 and 16 scene tokens, while keeping all other factors fixed. In both cases, the CFC-augmented model secures clear improvements over its non-cycle counterpart. Remarkably, even with only 8 tokens, our method can still learn the surrounding environment of the vehicle very well, confirming that the cycle constraint remains effective even in highly compact representations.

\textbf{Effect of the loss of weight.} As shown in Table~\ref {tab:4}, we show an ablation study on the weights of our three loss terms, with emphasis on the two components tied to the CFC BEV cycle. Keeping the trajectory loss $\mathcal{L} _{traj}$ fixed, we systematically vary the forward-pass loss $\mathcal{L} _{futbev}$ and the echo-pass loss $\mathcal{L} _{curbev}$. The results show that planning quality remains consistently high across a broad range of weight settings, underscoring the stability imparted by the CFC cycle and confirming that EchoP reliably strengthens planning performance.

\textbf{Why is the end-to-end CFC BEV cycle loop effective?} Accurate trajectory planning requires the model to form a faithful understanding of the driving scene. Most recent planners follow a one-shot paradigm: the network observes a single sensor frame, predicts the future trajectory, and is trained either with auxiliary labels such as detection or mapping or with a forward self-supervised loss that compares the predicted future scene to ground truth. This design leaves two critical gaps. First, heavy auxiliary heads greatly increase model complexity while offering no direct feedback on how well the planner actually understands its surroundings. Second, relying only on forward supervision ignores the reverse constraint: because the driving scene evolves unpredictably, a plan generated from the present frame alone may remain unverified and can drift from reality. EchoP addresses these issues with a CFC cycle. Without any extra supervision, the framework not only learns from the forward prediction of the future scene but also reconstructs the current BEV from that prediction, thereby checking whether the imagined future is consistent with the observed present. This bidirectional loop provides an intrinsic error detector that continuously validates trajectories and significantly lowers the probability of collisions, yielding more reliable planning.

\subsection{Qualitative Results}\label{sec:qualitative}
\textcolor{black}{\textbf{Visualization protocol.}
In Fig.~\ref{fig:3}and Fig.~\ref{fig:4}, map context is rendered from dataset annotations.
The green icon denotes the ego vehicle and red icons denote surrounding agents.
We overlay the predicted ego trajectory of EchoP with a representative one-shot baseline and the ground-truth trajectory to highlight behavioral differences under diverse conditions.}

\textbf{Visualization of planning trajectories.} In Figure~\ref{fig:3}, we qualitatively compare trajectories planned by EchoP with ground truth (GT) and GenAD~\cite{zheng2024genad} across four representative scenarios.
Figure~\ref{fig:3} (a) depicts an intersection requiring a left turn. EchoP generates a smoother and more anticipatory turning arc than the GT trajectory, aligning better with the lane geometry.
Figure~\ref{fig:3} (b) considers driving conditions under rain-induced poor visibility. Despite water puddles and reflections obscuring lane markings, EchoP accurately infers the roadway layout, generating a right-turn trajectory more faithful to the lane than the GT.
Figure~\ref{fig:3} (c) illustrates navigation through a complex multi-way junction. Here, EchoP closely follows the GT trajectory without crossing road boundaries, whereas GenAD drifts into non-drivable areas.
Figure~\ref{fig:3} (d) shows a scenario of low-light night driving. Under dim illumination, EchoP closely matches the GT trajectory, while GenAD exhibits significant deviations. These examples validate that the CFC cycle in EchoP intrinsically corrects trajectory predictions, enhancing robustness across diverse conditions.

\textbf{Failure cases.} In Figure~\ref{fig:4}, we show two representative failure modes.
The first stems from ambiguous navigation cues in the nuScenes annotation: in Fig.~\ref{fig:4} (a), the annotation marks the command as “go straight,” yet the ground-truth trajectory actually includes a rightward lane change. When such inconsistencies arise, the planner can be misled and produce a biased path.
The second limitation appears on very wide roads that lack explicit lane markings. Although the CFC cycle improves scene understanding, an open roadway grants considerable freedom even for human drivers. As a result, EchoP generates a trajectory that deviates from the ground truth, as in Fig.~\ref{fig:4} (b), while still remaining within a plausible driving corridor and preserving the correct turning direction.

\textcolor{black}{
\textbf{Real-world limitations and future directions.}
In very complex scenarios (e.g., weakly structured open areas, highly multi-modal interactions, and noisy command annotations), a single recorded reference may be insufficient to represent the set of safe and reasonable behaviors.
As a result, trajectory/waypoint regression to a single ground-truth target can over-penalize plausible alternatives, especially in open-space or interaction-heavy cases.
Future work may explore (i) improving supervision reliability (e.g., route-level command-consistency modeling), (ii) integrating optional geometric priors when available (e.g., drivable-area or road-boundary cues), and (iii) adopting multi-modal planning and corridor-aware evaluation to better reflect non-unique safe futures; deployment-level lightweight safety guardrails in closed-loop systems are also promising complements.}

\section{Conclusion}\label{5}
The Echo Planning framework introduces a self-correcting Current → Future → Current BEV cycle loop (CFC cycle) that endows end-to-end planners with an intrinsic notion of temporal consistency. By predicting a future trajectory, inverting it to reconstruct the present BEV, and penalising any mismatch through a cycle loss, the framework detects and suppresses implausible motion without extra labels or auxiliary heads. Extensive nuScenes and Bench2Drive experiments confirm that this bidirectional feedback reduces displacement error and collision rate beyond state-of-the-art one-shot planners, while incurring no annotation overhead and no additional inference overhead. Echo Planning therefore provides a practical, safety-oriented upgrade for modern autonomous-driving stacks. \textcolor{black}{Overall, this work highlights bidirectional trajectory--scene consistency as a deployable training principle for end-to-end planning, which may guide future robust planner design and benchmark diagnostics for complex scenarios.} We also note that extending this cycle-consistency principle to longer-horizon planning and other navigation settings (\eg, UAV navigation~\cite{chu2024towards}) remains an interesting direction for future work.

\noindent\textbf{Acknowledgment.} Z. Zheng acknowledges supports from Guangdong Basic and Applied Basic Research Foundation 2025A1515012281, the Jiangsu Provincial Science and Technology Program (Grant No. SBZ20250900116), the University of Macau Advanced Research Institute in Hengqin, and the Macao Science and Technology Development Fund Grant FDCT/0043/2025/RIA1.

\bibliographystyle{IEEEtran}
\bibliography{egbib}

@String(CVPR  = {CVPR})

@String(ICCV  = {ICCV})

@String(ECCV  = {ECCV})

@String(AAAI = {AAAI})

@String(CVPR= {IEEE Conf. Comput. Vis. Pattern Recog.})

@String(ICCV= {Int. Conf. Comput. Vis.})

@String(ECCV= {Eur. Conf. Comput. Vis.})

@ARTICLE{10791908,
  author={Li, Zhiqi and Wang, Wenhai and Li, Hongyang and Xie, Enze and Sima, Chonghao and Lu, Tong and Yu, Qiao and Dai, Jifeng},
  journal={IEEE Transactions on Pattern Analysis and Machine Intelligence}, 
  title={BEVFormer: Learning Bird’s-Eye-View Representation From LiDAR-Camera via Spatiotemporal Transformers}, 
  year={2025},
  volume={47},
  number={3},
  pages={2020-2036},
  doi={10.1109/TPAMI.2024.3515454}}

@inproceedings{linavigation,
  title={Navigation-Guided Sparse Scene Representation for End-to-End Autonomous Driving},
  author={Li, Peidong and Cui, Dixiao},
  year={2025},
  booktitle={The Thirteenth International Conference on Learning Representations}
}

@article{ryoo2021tokenlearner,
  title={Tokenlearner: Adaptive space-time tokenization for videos},
  author={Ryoo, Michael and Piergiovanni, AJ and Arnab, Anurag and Dehghani, Mostafa and Angelova, Anelia},
  journal={Advances in neural information processing systems},
  volume={34},
  pages={12786--12797},
  year={2021}
}

@article{ju2025video2bev,
  title={Video2bev: Transforming drone videos to bevs for video-based geo-localization},
  author={Ju, Hao and Huang, Shaofei and Liu, Si and Zheng, Zhedong},
  journal={ICCV},
  year={2025}
}

@inproceedings{Hu_Shen_Sun_2018,  
 title={Squeeze-and-Excitation Networks}, 
 url={http://dx.doi.org/10.1109/cvpr.2018.00745}, 
 DOI={10.1109/cvpr.2018.00745}, 
 booktitle={2018 IEEE/CVF Conference on Computer Vision and Pattern Recognition}, 
 author={Hu, Jie and Shen, Li and Sun, Gang}, 
 year={2018}, 
 month={Jun}, 
 language={en-US} 
 }

@article{Vaswani_Shazeer_Parmar_Uszkoreit_Jones_Gomez_Kaiser_Polosukhin_2017,  
 title={Attention is All you Need}, 
 journal={Neural Information Processing Systems,Neural Information Processing Systems}, 
 author={Vaswani, Ashish and Shazeer, Noam and Parmar, Niki and Uszkoreit, Jakob and Jones, Llion and Gomez, AidanN. and Kaiser, Lukasz and Polosukhin, Illia}, 
 year={2017}, 
 month={Jun}, 
 language={en-US} 
 }

@article{loshchilov2017decoupled,
  title={Decoupled weight decay regularization},
  author={Loshchilov, Ilya and Hutter, Frank},
  journal={arXiv preprint arXiv:1711.05101},
  year={2017}
}

@inproceedings{he2016deep,
  title={Deep residual learning for image recognition},
  author={He, Kaiming and Zhang, Xiangyu and Ren, Shaoqing and Sun, Jian},
  booktitle={Proceedings of the IEEE conference on computer vision and pattern recognition},
  pages={770--778},
  year={2016}
}

@inproceedings{jiang2023vad,
  title={Vad: Vectorized scene representation for efficient autonomous driving},
  author={Jiang, Bo and Chen, Shaoyu and Xu, Qing and Liao, Bencheng and Chen, Jiajie and Zhou, Helong and Zhang, Qian and Liu, Wenyu and Huang, Chang and Wang, Xinggang},
  booktitle={Proceedings of the IEEE/CVF International Conference on Computer Vision},
  pages={8340--8350},
  year={2023}
}

@inproceedings{hu2023planning,
  title={Planning-oriented autonomous driving},
  author={Hu, Yihan and Yang, Jiazhi and Chen, Li and Li, Keyu and Sima, Chonghao and Zhu, Xizhou and Chai, Siqi and Du, Senyao and Lin, Tianwei and Wang, Wenhai and others},
  booktitle={Proceedings of the IEEE/CVF conference on computer vision and pattern recognition},
  pages={17853--17862},
  year={2023}
}

@inproceedings{caesar2020nuscenes,
  title={nuscenes: A multimodal dataset for autonomous driving},
  author={Caesar, Holger and Bankiti, Varun and Lang, Alex H and Vora, Sourabh and Liong, Venice Erin and Xu, Qiang and Krishnan, Anush and Pan, Yu and Baldan, Giancarlo and Beijbom, Oscar},
  booktitle={Proceedings of the IEEE/CVF conference on computer vision and pattern recognition},
  pages={11621--11631},
  year={2020}
}

@inproceedings{zeng2019end,
  title={End-to-end interpretable neural motion planner},
  author={Zeng, Wenyuan and Luo, Wenjie and Suo, Simon and Sadat, Abbas and Yang, Bin and Casas, Sergio and Urtasun, Raquel},
  booktitle={Proceedings of the IEEE/CVF conference on computer vision and pattern recognition},
  pages={8660--8669},
  year={2019}
}

@inproceedings{hu2021safe,
  title={Safe local motion planning with self-supervised freespace forecasting},
  author={Hu, Peiyun and Huang, Aaron and Dolan, John and Held, David and Ramanan, Deva},
  booktitle={Proceedings of the IEEE/CVF Conference on Computer Vision and Pattern Recognition},
  pages={12732--12741},
  year={2021}
}

@inproceedings{khurana2022differentiable,
  title={Differentiable raycasting for self-supervised occupancy forecasting},
  author={Khurana, Tarasha and Hu, Peiyun and Dave, Achal and Ziglar, Jason and Held, David and Ramanan, Deva},
  booktitle={European Conference on Computer Vision},
  pages={353--369},
  year={2022},
  organization={Springer}
}

@inproceedings{hu2022st,
  title={St-p3: End-to-end vision-based autonomous driving via spatial-temporal feature learning},
  author={Hu, Shengchao and Chen, Li and Wu, Penghao and Li, Hongyang and Yan, Junchi and Tao, Dacheng},
  booktitle={European Conference on Computer Vision},
  pages={533--549},
  year={2022},
  organization={Springer}
}

@inproceedings{tong2023scene,
  title={Scene as occupancy},
  author={Tong, Wenwen and Sima, Chonghao and Wang, Tai and Chen, Li and Wu, Silei and Deng, Hanming and Gu, Yi and Lu, Lewei and Luo, Ping and Lin, Dahua and others},
  booktitle={Proceedings of the IEEE/CVF International Conference on Computer Vision},
  pages={8406--8415},
  year={2023}
}

@inproceedings{weng2024drive,
  title={Para-drive: Parallelized architecture for real-time autonomous driving},
  author={Weng, Xinshuo and Ivanovic, Boris and Wang, Yan and Wang, Yue and Pavone, Marco},
  booktitle={Proceedings of the IEEE/CVF Conference on Computer Vision and Pattern Recognition},
  pages={15449--15458},
  year={2024}
}

@inproceedings{zheng2024genad,
  title={Genad: Generative end-to-end autonomous driving},
  author={Zheng, Wenzhao and Song, Ruiqi and Guo, Xianda and Zhang, Chenming and Chen, Long},
  booktitle={European Conference on Computer Vision},
  pages={87--104},
  year={2024},
  organization={Springer}
}

@article{guo2024end,
  title={End-to-end autonomous driving without costly modularization and 3d manual annotation},
  author={Guo, Mingzhe and Zhang, Zhipeng and He, Yuan and Wang, Ke and Jing, Liping},
  journal={arXiv preprint arXiv:2406.17680},
  year={2024}
}

@inproceedings{li2024ego,
  title={Is ego status all you need for open-loop end-to-end autonomous driving?},
  author={Li, Zhiqi and Yu, Zhiding and Lan, Shiyi and Li, Jiahan and Kautz, Jan and Lu, Tong and Alvarez, Jose M},
  booktitle={Proceedings of the IEEE/CVF Conference on Computer Vision and Pattern Recognition},
  pages={14864--14873},
  year={2024}
}

@article{li2024enhancing,
  title={Enhancing end-to-end autonomous driving with latent world model},
  author={Li, Yingyan and Fan, Lue and He, Jiawei and Wang, Yuqi and Chen, Yuntao and Zhang, Zhaoxiang and Tan, Tieniu},
  journal={arXiv preprint arXiv:2406.08481},
  year={2024}
}

@article{sun2024sparsedrive,
  title={Sparsedrive: End-to-end autonomous driving via sparse scene representation},
  author={Sun, Wenchao and Lin, Xuewu and Shi, Yining and Zhang, Chuang and Wu, Haoran and Zheng, Sifa},
  journal={arXiv preprint arXiv:2405.19620},
  year={2024}
}

@article{Pomerleau_1988,  
 title={ALVINN: An Autonomous Land Vehicle in a Neural Network}, 
 DOI={10.1184/r1/6603146.v1}, 
 journal={Neural Information Processing Systems,Neural Information Processing Systems}, 
 author={Pomerleau, Dean}, 
 year={1988}, 
 month={Jan}, 
 language={en-US} 
 }

@article{Bojarski_Testa_Dworakowski_Firner_Flepp_Goyal_Jackel_Monfort_Muller_Zhang_etal, 
 title={End to End Learning for Self-Driving Cars}, 
 journal={arXiv: Computer Vision and Pattern Recognition,arXiv: Computer Vision and Pattern Recognition}, 
 author={Bojarski, Mariusz and Testa, DavideDel and Dworakowski, Daniel and Firner, Bernhard and Flepp, Beat and Goyal, Prasoon and Jackel, L.D. and Monfort, Mathew and Muller, Urs and Zhang, Jiakai and Zhang, Xin and Zhao, Jake and Zieba, Karol}, 
 year={2016}, 
 month={Apr}, 
 language={en-US} 
 }

@inproceedings{codevilla2019exploring,
  title={Exploring the limitations of behavior cloning for autonomous driving},
  author={Codevilla, Felipe and Santana, Eder and L{\'o}pez, Antonio M and Gaidon, Adrien},
  booktitle={Proceedings of the IEEE/CVF international conference on computer vision},
  pages={9329--9338},
  year={2019}
}

@article{chen2022efficient,
  title={Efficient and robust 2d-to-bev representation learning via geometry-guided kernel transformer},
  author={Chen, Shaoyu and Cheng, Tianheng and Wang, Xinggang and Meng, Wenming and Zhang, Qian and Liu, Wenyu},
  journal={arXiv preprint arXiv:2206.04584},
  year={2022}
}

@inproceedings{Hu_Murez_Mohan_Dudas_Hawke_Badrinarayanan_Cipolla_Kendall_2021,  
 title={FIERY: Future Instance Prediction in Bird’s-Eye View from Surround Monocular Cameras}, 
 url={http://dx.doi.org/10.1109/iccv48922.2021.01499}, 
 DOI={10.1109/iccv48922.2021.01499}, 
 booktitle={2021 IEEE/CVF International Conference on Computer Vision (ICCV)}, 
 author={Hu, Anthony and Murez, Zak and Mohan, Nikhil and Dudas, Sofia and Hawke, Jeffrey and Badrinarayanan, Vijay and Cipolla, Roberto and Kendall, Alex}, 
 year={2021}, 
 month={Oct}, 
 language={en-US} 
 }

@inproceedings{zhang2024approaching,
  title={Approaching outside: Scaling unsupervised 3d object detection from 2d scene},
  author={Zhang, Ruiyang and Zhang, Hu and Yu, Hang and Zheng, Zhedong},
  booktitle={European Conference on Computer Vision},
  pages={249--266},
  year={2024},
  organization={Springer}
}

@inproceedings{zhang2025harnessing,
  title={Harnessing uncertainty-aware bounding boxes for unsupervised 3d object detection},
  author={Zhang, Ruiyang and Zhang, Hu and Zheng, Zhedong},
  booktitle={Proceedings of the IEEE/CVF International Conference on Computer Vision},
  pages={9230--9240},
  year={2025}
}

@inproceedings{chu2024towards,
  title={Towards natural language-guided drones: GeoText-1652 benchmark with spatial relation matching},
  author={Chu, Meng and Zheng, Zhedong and Ji, Wei and Wang, Tingyu and Chua, Tat-Seng},
  booktitle={European Conference on Computer Vision},
  pages={213--231},
  year={2024},
  organization={Springer}
}

@inproceedings{liao2024lane,
  title={Lane graph as path: Continuity-preserving path-wise modeling for online lane graph construction},
  author={Liao, Bencheng and Chen, Shaoyu and Jiang, Bo and Cheng, Tianheng and Zhang, Qian and Liu, Wenyu and Huang, Chang and Wang, Xinggang},
  booktitle={European Conference on Computer Vision},
  pages={334--351},
  year={2024},
  organization={Springer}
}

@article{suo2024jointly,
  title={Jointly harnessing prior structures and temporal consistency for sign language video generation},
  author={Suo, Yucheng and Zheng, Zhedong and Wang, Xiaohan and Zhang, Bang and Yang, Yi},
  journal={ACM Transactions on Multimedia Computing, Communications and Applications},
  volume={20},
  number={6},
  pages={1--18},
  year={2024},
  publisher={ACM New York, NY}
}

@article{liao2022maptr,
  title={Maptr: Structured modeling and learning for online vectorized hd map construction},
  author={Liao, Bencheng and Chen, Shaoyu and Wang, Xinggang and Cheng, Tianheng and Zhang, Qian and Liu, Wenyu and Huang, Chang},
  journal={arXiv preprint arXiv:2208.14437},
  year={2022}
}

@inproceedings{liu2023vision,
  title={Vision-based uneven bev representation learning with polar rasterization and surface estimation},
  author={Liu, Zhi and Chen, Shaoyu and Guo, Xiaojie and Wang, Xinggang and Cheng, Tianheng and Zhu, Hongmei and Zhang, Qian and Liu, Wenyu and Zhang, Yi},
  booktitle={Conference on Robot Learning},
  pages={437--446},
  year={2023},
  organization={PMLR}
}

@article{zhang2022beverse,
  title={Beverse: Unified perception and prediction in birds-eye-view for vision-centric autonomous driving},
  author={Zhang, Yunpeng and Zhu, Zheng and Zheng, Wenzhao and Huang, Junjie and Huang, Guan and Zhou, Jie and Lu, Jiwen},
  journal={arXiv preprint arXiv:2205.09743},
  year={2022}
}

@inbook{Philion_Fidler_2020,  
 title={Lift, Splat, Shoot: Encoding Images from Arbitrary Camera Rigs by Implicitly Unprojecting to 3D}, 
 url={http://dx.doi.org/10.1007/978-3-030-58568-6_12}, 
 DOI={10.1007/978-3-030-58568-6_12}, 
 booktitle={ECCV}, 
 author={Philion, Jonah and Fidler, Sanja}, 
 year={2020}, 
 month={Jan}, 
 pages={194–210}, 
 language={en-US} 
 }

@inproceedings{chen2023voxelnext,
  title={Voxelnext: Fully sparse voxelnet for 3d object detection and tracking},
  author={Chen, Yukang and Liu, Jianhui and Zhang, Xiangyu and Qi, Xiaojuan and Jia, Jiaya},
  booktitle={Proceedings of the IEEE/CVF Conference on Computer Vision and Pattern Recognition},
  pages={21674--21683},
  year={2023}
}

@inproceedings{Graham_Engelcke_Maaten_2018,  
 title={3D Semantic Segmentation with Submanifold Sparse Convolutional Networks}, 
 url={http://dx.doi.org/10.1109/cvpr.2018.00961}, 
 DOI={10.1109/cvpr.2018.00961}, 
 booktitle={2018 IEEE/CVF Conference on Computer Vision and Pattern Recognition}, 
 author={Graham, Benjamin and Engelcke, Martin and Maaten, Laurens van der}, 
 year={2018}, 
 month={Jun}, 
 language={en-US} 
 }

@inproceedings{Mao_Xue_Niu_Bai_Feng_Liang_Xu_Xu_2021,  
 title={Voxel Transformer for 3D Object Detection}, 
 url={http://dx.doi.org/10.1109/iccv48922.2021.00315}, 
 DOI={10.1109/iccv48922.2021.00315}, 
 booktitle={2021 IEEE/CVF International Conference on Computer Vision (ICCV)}, 
 author={Mao, Jiageng and Xue, Yujing and Niu, Minzhe and Bai, Haoyue and Feng, Jiashi and Liang, Xiaodan and Xu, Hang and Xu, Chunjing}, 
 year={2021}, 
 month={Oct}, 
 language={en-US} 
 }

@article{zhang2024sparsead,
  title={Sparsead: Sparse query-centric paradigm for efficient end-to-end autonomous driving},
  author={Zhang, Diankun and Wang, Guoan and Zhu, Runwen and Zhao, Jianbo and Chen, Xiwu and Zhang, Siyu and Gong, Jiahao and Zhou, Qibin and Zhang, Wenyuan and Wang, Ningzi and others},
  journal={arXiv preprint arXiv:2404.06892},
  year={2024}
}

@inproceedings{chen2024ppad,
  title={Ppad: Iterative interactions of prediction and planning for end-to-end autonomous driving},
  author={Chen, Zhili and Ye, Maosheng and Xu, Shuangjie and Cao, Tongyi and Chen, Qifeng},
  booktitle={European Conference on Computer Vision},
  pages={239--256},
  year={2024},
  organization={Springer}
}

@inproceedings{wang2023exploring,
  title={Exploring object-centric temporal modeling for efficient multi-view 3d object detection},
  author={Wang, Shihao and Liu, Yingfei and Wang, Tiancai and Li, Ying and Zhang, Xiangyu},
  booktitle={Proceedings of the IEEE/CVF international conference on computer vision},
  pages={3621--3631},
  year={2023}
}

@inproceedings{Codevilla_Muller_Lopez_Koltun_Dosovitskiy_2018,   title={End-to-end Driving via Conditional Imitation Learning},  url={http://dx.doi.org/10.1109/icra.2018.8460487},  DOI={10.1109/icra.2018.8460487},  booktitle={2018 IEEE International Conference on Robotics and Automation (ICRA)},  author={Codevilla, Felipe and Muller, Matthias and Lopez, Antonio and Koltun, Vladlen and Dosovitskiy, Alexey},  year={2018},  month={May},  language={en-US}  }

@inproceedings{Prakash_Chitta_Geiger_2021,  
 title={Multi-Modal Fusion Transformer for End-to-End Autonomous Driving}, 
 url={http://dx.doi.org/10.1109/cvpr46437.2021.00700}, 
 DOI={10.1109/cvpr46437.2021.00700}, 
 booktitle={2021 IEEE/CVF Conference on Computer Vision and Pattern Recognition (CVPR)}, 
 author={Prakash, Aditya and Chitta, Kashyap and Geiger, Andreas}, 
 year={2021}, 
 month={Jun}, 
 language={en-US} 
 }

@article{zhu2024autonomous,
  title={Autonomous driving with spiking neural networks},
  author={Zhu, Rui-Jie and Wang, Ziqing and Gilpin, Leilani and Eshraghian, Jason},
  journal={Advances in Neural Information Processing Systems},
  volume={37},
  pages={136782--136804},
  year={2024}
}

@inproceedings{jia2024bench,
  title={Bench2Drive: Towards Multi-Ability Benchmarking of Closed-Loop End-To-End Autonomous Driving},
  author={Xiaosong Jia and Zhenjie Yang and Qifeng Li and Zhiyuan Zhang and Junchi Yan},
  booktitle={NeurIPS 2024 Datasets and Benchmarks Track},
  year={2024}
}

@article{ye2023fusionad,
  title={Fusionad: Multi-modality fusion for prediction and planning tasks of autonomous driving},
  author={Ye, Tengju and Jing, Wei and Hu, Chunyong and Huang, Shikun and Gao, Lingping and Li, Fangzhen and Wang, Jingke and Guo, Ke and Xiao, Wencong and Mao, Weibo and others},
  journal={arXiv preprint arXiv:2308.01006},
  year={2023}
}

@article{zhang2024graphad,
  title={Graphad: Interaction scene graph for end-to-end autonomous driving},
  author={Zhang, Yunpeng and Qian, Deheng and Li, Ding and Pan, Yifeng and Chen, Yong and Liang, Zhenbao and Zhang, Zhiyao and Zhang, Shurui and Li, Hongxu and Fu, Maolei and others},
  journal={arXiv preprint arXiv:2403.19098},
  year={2024}
}

@article{huang2021bevdet,
  title={Bevdet: High-performance multi-camera 3d object detection in bird-eye-view},
  author={Huang, Junjie and Huang, Guan and Zhu, Zheng and Ye, Yun and Du, Dalong},
  journal={arXiv preprint arXiv:2112.11790},
  year={2021}
}

@inproceedings{li2023bevdepth,
  title={Bevdepth: Acquisition of reliable depth for multi-view 3d object detection},
  author={Li, Yinhao and Ge, Zheng and Yu, Guanyi and Yang, Jinrong and Wang, Zengran and Shi, Yukang and Sun, Jianjian and Li, Zeming},
  booktitle={Proceedings of the AAAI conference on artificial intelligence},
  volume={37},
  number={2},
  pages={1477--1485},
  year={2023}
}

@inproceedings{dauner2023parting,
  title={Parting with misconceptions about learning-based vehicle motion planning},
  author={Dauner, Daniel and Hallgarten, Marcel and Geiger, Andreas and Chitta, Kashyap},
  booktitle={Conference on Robot Learning},
  pages={1268--1281},
  year={2023},
  organization={PMLR}
}

@inproceedings{cheng2022mpnp,
  title={Mpnp: Multi-policy neural planner for urban driving},
  author={Cheng, Jie and Xin, Ren and Wang, Sheng and Liu, Ming},
  booktitle={2022 IEEE/RSJ International Conference on Intelligent Robots and Systems (IROS)},
  pages={10549--10554},
  year={2022},
  organization={IEEE}
}

@article{wu2022trajectory,
  title={Trajectory-guided control prediction for end-to-end autonomous driving: A simple yet strong baseline},
  author={Wu, Penghao and Jia, Xiaosong and Chen, Li and Yan, Junchi and Li, Hongyang and Qiao, Yu},
  journal={Advances in Neural Information Processing Systems},
  volume={35},
  pages={6119--6132},
  year={2022}
}

@inproceedings{jia2023think,
  title={Think twice before driving: Towards scalable decoders for end-to-end autonomous driving},
  author={Jia, Xiaosong and Wu, Penghao and Chen, Li and Xie, Jiangwei and He, Conghui and Yan, Junchi and Li, Hongyang},
  booktitle={Proceedings of the IEEE/CVF Conference on Computer Vision and Pattern Recognition},
  pages={21983--21994},
  year={2023}
}

@inproceedings{liu2023bevfusion,
  title={Bevfusion: Multi-task multi-sensor fusion with unified bird's-eye view representation},
  author={Liu, Zhijian and Tang, Haotian and Amini, Alexander and Yang, Xinyu and Mao, Huizi and Rus, Daniela L and Han, Song},
  booktitle={2023 IEEE international conference on robotics and automation (ICRA)},
  pages={2774--2781},
  year={2023},
  organization={IEEE}
}

@inproceedings{shao2023reasonnet,
  title={Reasonnet: End-to-end driving with temporal and global reasoning},
  author={Shao, Hao and Wang, Letian and Chen, Ruobing and Waslander, Steven L and Li, Hongsheng and Liu, Yu},
  booktitle={Proceedings of the IEEE/CVF conference on computer vision and pattern recognition},
  pages={13723--13733},
  year={2023}
}

@article{yuan2024drama,
  title={Drama: An efficient end-to-end motion planner for autonomous driving with mamba},
  author={Yuan, Chengran and Zhang, Zhanqi and Sun, Jiawei and Sun, Shuo and Huang, Zefan and Lee, Christina Dao Wen and Li, Dongen and Han, Yuhang and Wong, Anthony and Tee, Keng Peng and others},
  journal={arXiv preprint arXiv:2408.03601},
  year={2024}
}

@misc{zhai2023rethinkingopenloopevaluationendtoend,
      title={Rethinking the Open-Loop Evaluation of End-to-End Autonomous Driving in nuScenes}, 
      author={Jiang-Tian Zhai and Ze Feng and Jinhao Du and Yongqiang Mao and Jiang-Jiang Liu and Zichang Tan and Yifu Zhang and Xiaoqing Ye and Jingdong Wang},
      year={2023},
      eprint={2305.10430},
      archivePrefix={arXiv},
      primaryClass={cs.CV},
      url={https://arxiv.org/abs/2305.10430}, 
}

@article{Liao2024DiffusionDriveTD,
  title={DiffusionDrive: Truncated Diffusion Model for End-to-End Autonomous Driving},
  author={Bencheng Liao and Shaoyu Chen and Haoran Yin and Bo Jiang and Cheng Wang and Sixu Yan and Xinbang Zhang and Xiangyu Li and Ying Zhang and Qian Zhang and Xinggang Wang},
  journal={2025 IEEE/CVF Conference on Computer Vision and Pattern Recognition (CVPR)},
  year={2024},
  pages={12037-12047},
  url={https://api.semanticscholar.org/CorpusID:274192736}
}

@article{wang2024he,
  title={He-drive: Human-like end-to-end driving with vision language models},
  author={Wang, Junming and Zhang, Xingyu and Xing, Zebin and Gu, Songen and Guo, Xiaoyang and Hu, Yang and Song, Ziying and Zhang, Qian and Long, Xiaoxiao and Yin, Wei},
  journal={arXiv preprint arXiv:2410.05051},
  year={2024}
}

@inproceedings{song2025don,
  title={Don't Shake the Wheel: Momentum-Aware Planning in End-to-End Autonomous Driving},
  author={Song, Ziying and Jia, Caiyan and Liu, Lin and Pan, Hongyu and Zhang, Yongchang and Wang, Junming and Zhang, Xingyu and Xu, Shaoqing and Yang, Lei and Luo, Yadan},
  booktitle={Proceedings of the Computer Vision and Pattern Recognition Conference},
  pages={22432--22441},
  year={2025}
}

@inproceedings{zhang2025bridging,
  title={Bridging past and future: End-to-end autonomous driving with historical prediction and planning},
  author={Zhang, Bozhou and Song, Nan and Jin, Xin and Zhang, Li},
  booktitle={Proceedings of the Computer Vision and Pattern Recognition Conference},
  pages={6854--6863},
  year={2025}
}

@article{10.1109/TPAMI.2024.3483273,
author = {Wang, Wenguan and Yang, Yi and Wu, Fei},
title = {Towards Data-And Knowledge-Driven AI: A Survey on Neuro-Symbolic Computing},
year = {2025},
issue_date = {Feb. 2025},
publisher = {IEEE Computer Society},
address = {USA},
volume = {47},
number = {2},
issn = {0162-8828},
url = {https://doi.org/10.1109/TPAMI.2024.3483273},
doi = {10.1109/TPAMI.2024.3483273},
journal = {IEEE Trans. Pattern Anal. Mach. Intell.},
month = feb,
pages = {878–899},
numpages = {22}
}

@article{feng2025survey,
  title={A survey of world models for autonomous driving},
  author={Feng, Tuo and Wang, Wenguan and Yang, Yi},
  journal={arXiv preprint arXiv:2501.11260},
  year={2025}
}

@article{10.1109/TPAMI.2024.3508798,
author = {Huang, Shaofei and Shen, Zhenwei and Huang, Zehao and Liao, Yue and Han, Jizhong and Wang, Naiyan and Liu, Si},
title = {Anchor3DLane++: 3D Lane Detection via Sample-Adaptive Sparse 3D Anchor Regression},
year = {2025},
issue_date = {March 2025},
publisher = {IEEE Computer Society},
address = {USA},
volume = {47},
number = {3},
issn = {0162-8828},
url = {https://doi.org/10.1109/TPAMI.2024.3508798},
doi = {10.1109/TPAMI.2024.3508798},
journal = {IEEE Trans. Pattern Anal. Mach. Intell.},
month = mar,
pages = {1660–1673},
numpages = {14}
}

@ARTICLE{11175545,
  author={Lu, Jiachen and Nie, Ming and Zhang, Bozhou and Peng, Renyuan and Cai, Xinyue and Xu, Hang and Wen, Feng and Zhang, Wei and Zhang, Li},
  journal={IEEE Transactions on Pattern Analysis and Machine Intelligence}, 
  title={Translating Images to Road Network: A Sequence-to-Sequence Perspective}, 
  year={2025},
  volume={},
  number={},
  pages={1-18},
  doi={10.1109/TPAMI.2025.3612940}}

@ARTICLE{9729103,
  author={Ji, Wei and Yan, Ge and Li, Jingjing and Piao, Yongri and Yao, Shunyu and Zhang, Miao and Cheng, Li and Lu, Huchuan},
  journal={IEEE Transactions on Image Processing}, 
  title={DMRA: Depth-Induced Multi-Scale Recurrent Attention Network for RGB-D Saliency Detection}, 
  year={2022},
  volume={31},
  number={},
  pages={2321-2336},
  doi={10.1109/TIP.2022.3154931}}

@ARTICLE{10980037,
  author={Yang, Zeyu and Song, Nan and Li, Wei and Zhu, Xiatian and Zhang, Li and Torr, Philip H.S.},
  journal={IEEE Transactions on Pattern Analysis and Machine Intelligence}, 
  title={DeepInteraction++: Multi-Modality Interaction for Autonomous Driving}, 
  year={2025},
  volume={47},
  number={8},
  pages={6749-6763},
  doi={10.1109/TPAMI.2025.3565194}}

@article{yan2019stat,
  title={STAT: Spatial-temporal attention mechanism for video captioning},
  author={Yan, Chenggang and Tu, Yunbin and Wang, Xingzheng and Zhang, Yongbing and Hao, Xinyu and Zhang, Yu and Dai, Qionghai},
  journal={IEEE Transactions on Multimedia},
  volume={22},
  number={1},
  pages={229--241},
  year={2019},
  publisher={IEEE}
}

@misc{minddrive2025,
  title         = {MindDrive: An All-in-One Framework Bridging World Models and Vision-Language Model for End-to-End Autonomous Driving},
  author        = {Bin Sun and Yaoguang Cao and Yan Wang and Rui Wang and Jiachen Shang and Xiejie Feng and Jiayi Lu and Jia Shi and Shichun Yang and Xiaoyu Yan and Ziying Song},
  year          = {2025},
  eprint        = {2512.04441},
  archivePrefix = {arXiv},
  primaryClass  = {cs.CV},
  doi           = {10.48550/arXiv.2512.04441}
}

@misc{guideflow2025,
  title         = {GuideFlow: Constraint-Guided Flow Matching for Planning in End-to-End Autonomous Driving},
  author        = {Lin Liu and Caiyan Jia and Guanyi Yu and Ziying Song and JunQiao Li and Feiyang Jia and Peiliang Wu and Xiaoshuai Hao and Yandan Luo},
  year          = {2025},
  eprint        = {2511.18729},
  archivePrefix = {arXiv},
  primaryClass  = {cs.CV},
  doi           = {10.48550/arXiv.2511.18729}
}

@misc{diver2025,
  title         = {DIVER: Reinforced Diffusion Breaks Imitation Bottlenecks in End-to-End Autonomous Driving},
  author        = {Ziying Song and Lin Liu and Hongyu Pan and Bencheng Liao and Mingzhe Guo and Lei Yang and Yongchang Zhang and Shaoqing Xu and Caiyan Jia and Yadan Luo},
  year          = {2025},
  eprint        = {2507.04049},
  archivePrefix = {arXiv},
  primaryClass  = {cs.CV},
  doi           = {10.48550/arXiv.2507.04049}
}

@misc{focalad2025,
  title         = {FocalAD: Local Motion Planning for End-to-End Autonomous Driving},
  author        = {Bin Sun and Boao Zhang and Jiayi Lu and Xinjie Feng and Jiachen Shang and Rui Cao and Mengchao Zheng and Chuanye Wang and Shichun Yang and Yaoguang Cao and Ziying Song},
  year          = {2025},
  eprint        = {2506.11419},
  archivePrefix = {arXiv},
  primaryClass  = {cs.AI},
  doi           = {10.48550/arXiv.2506.11419}
}

@misc{vadv22024,
  title         = {VADv2: End-to-End Vectorized Autonomous Driving via Probabilistic Planning},
  author        = {Shaoyu Chen and Bo Jiang and Hao Gao and Bencheng Liao and Qing Xu and Qian Zhang and Chang Huang and Wenyu Liu and Xinggang Wang},
  year          = {2024},
  eprint        = {2402.13243},
  archivePrefix = {arXiv},
  primaryClass  = {cs.CV},
  doi           = {10.48550/arXiv.2402.13243}
}
\vspace{-10mm} 
\begin{IEEEbiography}[{\includegraphics[width=1in,height=1in,clip,keepaspectratio]{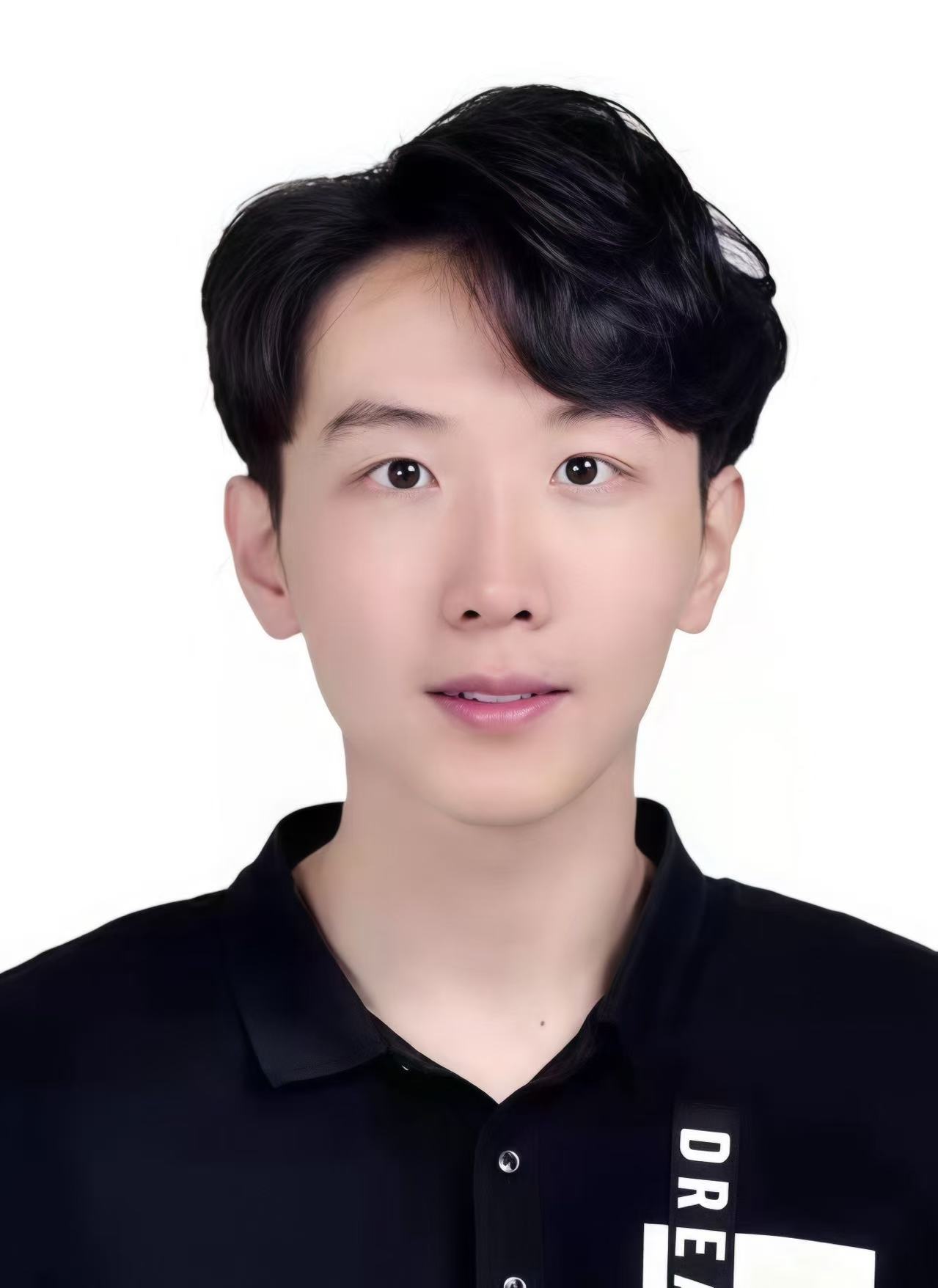}}]{Jintao Sun} received the M.S. degree from The George Washington University, America, in 2021 and the B.E. degree from the Harbin Institute of Technology, China, in 2019. He is currently working toward the Ph.D. degree with the School of Computer Science and Technology, Beijing Institute of Technology, China. His current research interests include the image retrieval, vision and language and the autonomous driving. 
\end{IEEEbiography}
\vspace{-12mm} 
\begin{IEEEbiography}[{\includegraphics[width=1in,height=1in,clip,keepaspectratio]{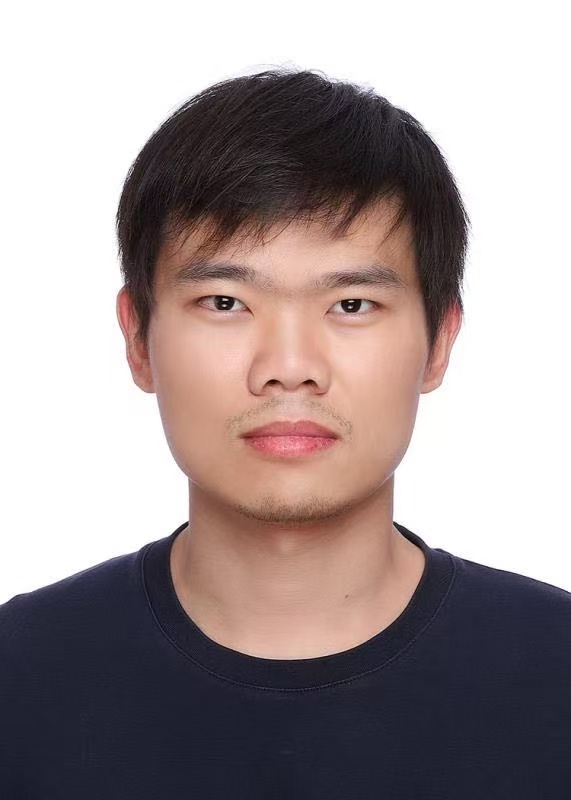}}]{Hu Zhang} is a CERC Research Fellow in CSIRO DATA61, Australia. He received the Ph.D. degree from the University of Technology Sydney in 2022 and the B.S. degree from University of Science and Technology of China in 2017. His research interests include 3D computer vision in Autonomous Driving, 3D reconstruction, imbalanced data learning.
\end{IEEEbiography}
\vspace{-15mm}
\begin{IEEEbiography}[{\includegraphics[width=1in,height=1in,clip,keepaspectratio]{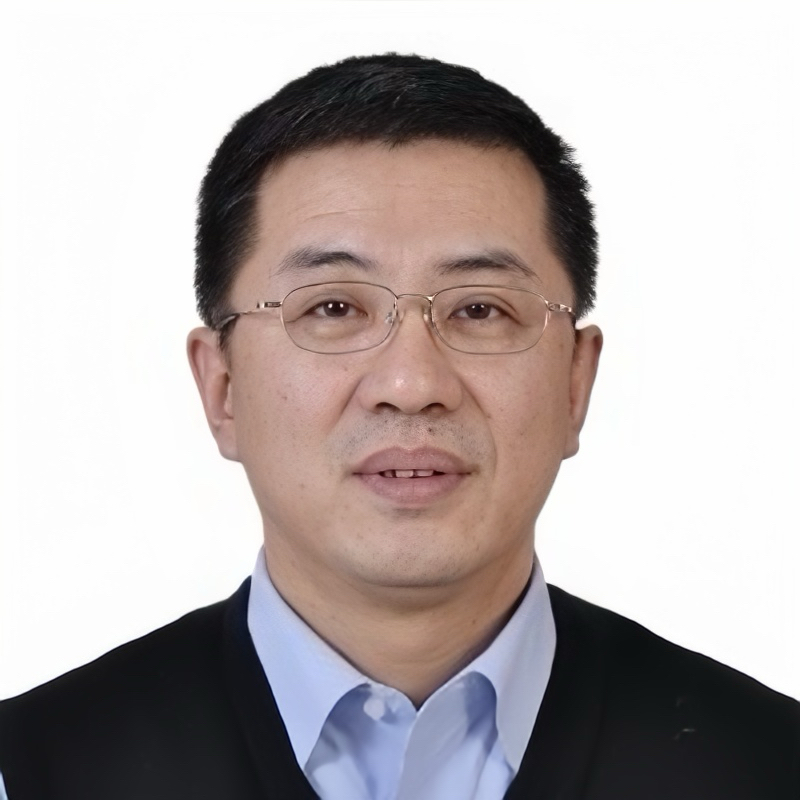}}]{Gangyi Ding} received the B.E. degree from Peking
University, Beijing, China, in 1988 and the Ph.D.
degree from the Beijing Institute of Technology, Beijing, in 1993. He is currently a Professor with the School of Computer Science and Technology, Beijing
Institute of Technology. In 1993, he joined the faculty,
Beijing Institute of Technology. His research interests
include computer simulation, software engineering,
and digital performance.
\end{IEEEbiography}
\vspace{-12mm}
\begin{IEEEbiography}[{\includegraphics[width=1in,height=1in,clip,keepaspectratio]{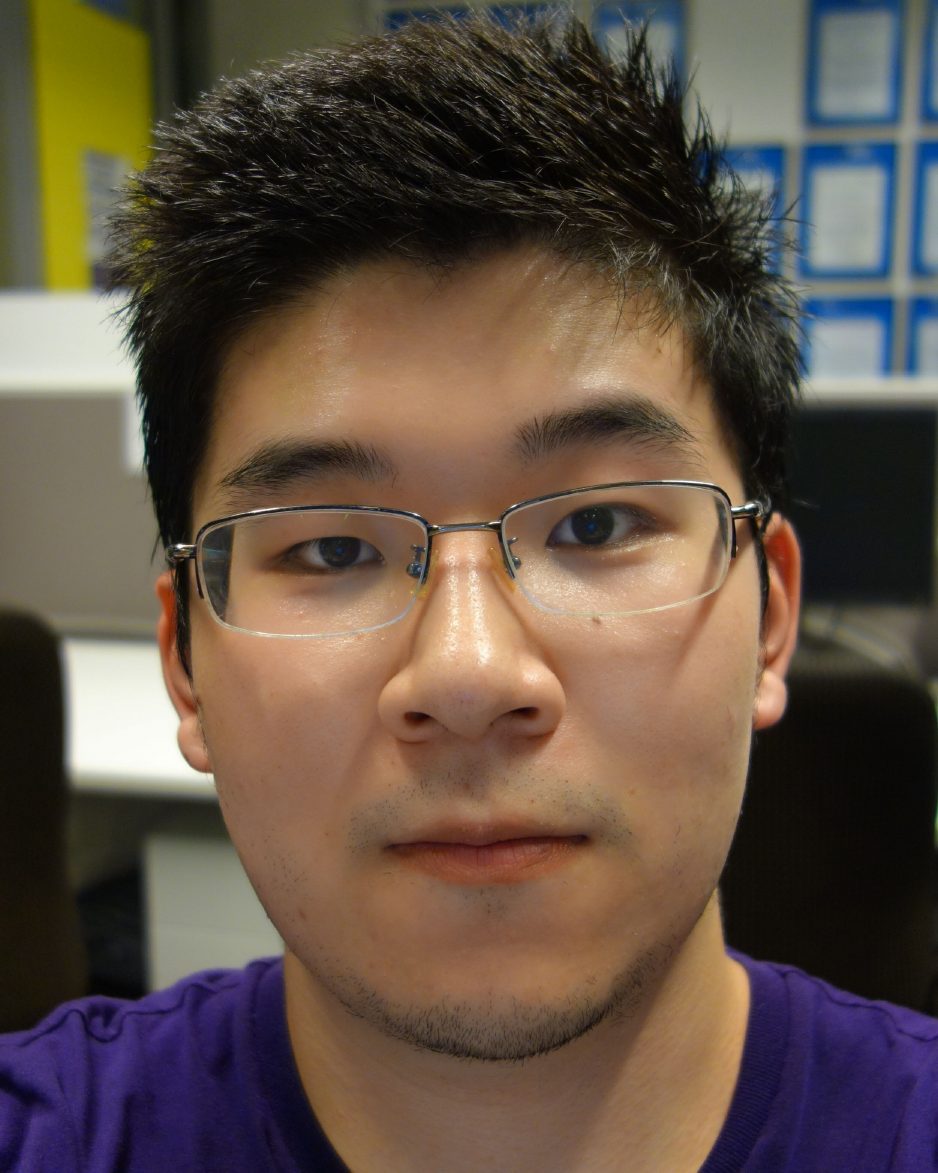}}]{Zhedong Zheng} is an Assistant Professor with the University of Macau. 
He received the Ph.D. degree from the University of Technology Sydney in 2021 and the B.S. degree from Fudan University in 2016. He was a postdoctoral research fellow at the School of Computing, National University of Singapore. He received the IEEE Circuits and Systems Society Outstanding Young Author Award of 2021. 
His research interests include robust learning for image retrieval, generative learning for data augmentation, and unsupervised domain adaptation. He served as the senior PC for IJCAI and AAAI, and the area chair for ACM MM'24 and ICASSP'25.
\end{IEEEbiography}

\vfill

\end{document}